\PassOptionsToPackage{unicode}{hyperref}
\PassOptionsToPackage{hyphens}{url}
\PassOptionsToPackage{dvipsnames,svgnames,x11names}{xcolor}
\documentclass{article}
\usepackage{iclr2027_conference}
\usepackage{times}
\usepackage{booktabs}
\usepackage{graphicx}
\usepackage{multirow}
\usepackage{xcolor}
\usepackage{colortbl}
\usepackage{caption}
\usepackage{float}
\usepackage{tabularx}
\usepackage{amsmath,amssymb}
\usepackage{iftex}
\ifPDFTeX
  \usepackage[T1]{fontenc}
  \usepackage[utf8]{inputenc}
  \usepackage{textcomp} 
\else 
  \usepackage{unicode-math} 
  \defaultfontfeatures{Scale=MatchLowercase}
  \defaultfontfeatures[\rmfamily]{Ligatures=TeX,Scale=1}
\fi
\ifPDFTeX\else
\fi
\IfFileExists{upquote.sty}{\usepackage{upquote}}{}
\IfFileExists{microtype.sty}{
  \usepackage[]{microtype}
  \UseMicrotypeSet[protrusion]{basicmath} 
}{}
\usepackage{longtable,array}
\usepackage{calc} 
\usepackage{etoolbox}
\makeatletter
\patchcmd\longtable{\par}{\if@noskipsec\mbox{}\fi\par}{}{}
\makeatother
\IfFileExists{footnotehyper.sty}{\usepackage{footnotehyper}}{\usepackage{footnote}}
\makesavenoteenv{longtable}
\makeatletter
\newsavebox\pandoc@box
\newcommand*\pandocbounded[1]{
  \sbox\pandoc@box{#1}%
  \Gscale@div\@tempa{\textheight}{\dimexpr\ht\pandoc@box+\dp\pandoc@box\relax}%
  \Gscale@div\@tempb{\linewidth}{\wd\pandoc@box}%
  \ifdim\@tempb\p@<\@tempa\p@\let\@tempa\@tempb\fi
  \ifdim\@tempa\p@<\p@\scalebox{\@tempa}{\usebox\pandoc@box}%
  \else\usebox{\pandoc@box}%
  \fi%
}
\def\fps@figure{htbp}
\makeatother
\providecommand{\tightlist}{%
  \setlength{\itemsep}{0pt}\setlength{\parskip}{0pt}}
\usepackage[]{natbib}
\usepackage{bookmark}
\IfFileExists{xurl.sty}{\usepackage{xurl}}{} 
\hypersetup{
  pdftitle={CoFlu-Brain: An fMRI Connectivity Foundation Model with Cofluctuation-Based Pretraining},
  pdfauthor={Ziteng Sui, Junfeng Xia, Wenhao Ye, Mo Wang, Quanying Liu},
  colorlinks=true,
  linkcolor={Maroon},
  filecolor={Maroon},
  citecolor={Blue},
  urlcolor={Blue},
  pdfcreator={LaTeX via pandoc}}

\definecolor{OmniRed}{RGB}{205,82,96}
\definecolor{TableTint}{RGB}{239,239,239}
\newcommand{\best}[1]{\textcolor{OmniRed}{\bfseries #1}}

\AtBeginEnvironment{table}{\renewcommand{\arraystretch}{1.08}}
\AtBeginEnvironment{longtable}{\renewcommand{\arraystretch}{1.08}}

\title{CoFlu-Brain: An fMRI Connectivity Foundation Model with Cofluctuation-Based Pretraining}
\iclrfinalcopy
\author{\normalfont\begin{minipage}{0.97\textwidth}
\centering
Ziteng Sui\textsuperscript{1}, Junfeng Xia\textsuperscript{1}, Wenhao Ye\textsuperscript{1,3},
Mo Wang\textsuperscript{1,2,*}, Quanying Liu\textsuperscript{1,*}\\[6pt]
{\small
\textsuperscript{1}Department of Biomedical Engineering,\\
Southern University of Science and Technology, China\\
\textsuperscript{2}Department of Computer Science, University of Warwick, UK\\
\textsuperscript{3}School of Biomedical Engineering, Shenzhen University, China\\[5pt]
\textsuperscript{*}Corresponding authors:\\
\href{mailto:12250099@mail.sustech.edu.cn}{12250099@mail.sustech.edu.cn},
\href{mailto:liuqy@sustech.edu.cn}{liuqy@sustech.edu.cn}}
\end{minipage}}
\date{}

\begin{document}
\maketitle
\lhead{Preprint}
\raggedbottom

\begin{abstract}

Functional magnetic resonance imaging (fMRI) commonly characterizes
large-scale brain organization through functional connectivity (FC). However,
an observed FC matrix is not a fixed brain graph but a statistical estimate
determined by the finite BOLD frames used to compute it. Conventional FC-based
approaches typically represent each scan with a single Pearson-correlation
matrix, leaving frame-resolved variations in regional interactions
unexploited. To address this limitation, we introduce CoFlu-Brain, which
organizes framewise cofluctuation observations into multiple retained and
dropped views, each summarized as an FC matrix. The model learns connectivity
structure reproducible across retained views and predicts the representations
of the corresponding dropped views, while adopting SIGReg from LeJEPA to
prevent representation collapse. We pretrained CoFlu-Brain on 13,444
subjects from eight dataset groups and evaluated its representations via
frozen-encoder linear probing on seven classification and two age-prediction
tasks across seven datasets.
CoFlu-Brain showed strong transfer across demographic, developmental, and
clinical prediction tasks, outperforming the evaluated released models on
most tasks. Random frame partitioning improved aggregate
transfer, while the complete objective achieved the strongest overall
performance.
\href{https://anonymous.4open.science/r/review-artifact-2027/}{Code is available.}

\end{abstract}

\section{Introduction}\label{introduction}

Functional magnetic resonance imaging (fMRI) provides a non-invasive view of
large-scale functional brain organization and its relationships with
cognition, behavior, and disease. As large multi-site fMRI collections
continue to grow, self-supervised pretraining offers a path toward
transferable brain representations without requiring downstream task labels
during pretraining \citep{brainlm,brainjepa,semantoks}. A common route from
fMRI measurements to brain-network representations is to parcellate the brain
into regions of interest (ROIs), extract a regional BOLD time series from each
ROI, and estimate the statistical dependence between regional signals.
Pearson correlation organizes these pairwise dependencies into a compact and
interpretable functional connectivity (FC) matrix. By explicitly representing
interregional functional coupling, FC has become a widely used input for
brain-network modeling and connectome-based pretraining
\citep{brainnettransformer,brainmass,braingfm}.

An observed Pearson-FC matrix, however, is not a sampling-independent brain
graph but a scan-level statistical estimate formed from a finite BOLD
sequence. Its value can vary with the frames included in the estimation, such
that different frame compositions produce non-identical but statistically
related FC estimates.

Existing FC representation models typically operate on a single Pearson-FC
matrix or graph for each scan \citep{brainnettransformer,braingfm}, leaving
frame-resolved variation unused during representation learning. Framewise
cofluctuation provides a finer-grained description of regional interactions by
characterizing the joint standardized activity of regional pairs at individual
frames \citep{esfahlani2020high,faskowitz2020edge}. This perspective makes it
possible to construct multiple training views from different compositions of
framewise observations.

This gap raises a central question: can different compositions of framewise
cofluctuation observations provide self-supervision for learning stable and
transferable connectivity representations? We hypothesize that retained frame
sets sampled from the same segment provide different views of subject-specific
cofluctuation structure. Although the resulting FC views may differ
numerically, their shared connectivity structure should remain reproducibly
inferable. Moreover, a retained view should contain sufficient information to
predict the representation of the corresponding non-overlapping dropped view.
The former encourages consistency across retained views, whereas the latter
captures predictive relationships between retained and dropped observations.

Motivated by this gap, we propose \textbf{CoFlu-Brain}, a
cofluctuation-based approach to self-supervised FC pretraining. For each fMRI
segment, CoFlu-Brain repeatedly partitions the frames into retained and
dropped sets, organizes their framewise cofluctuation observations into
separate FC views, and uses the dropped-view representations as prediction
targets. Its joint-embedding objective promotes consistent representations
across retained views while requiring each retained representation to predict
its corresponding dropped-view representation in latent space. We evaluate
the resulting representations through multi-dataset pretraining, frozen
transfer, and systematic ablations.

Our contributions are:

\begin{enumerate}
\def\labelenumi{\arabic{enumi}.}
\tightlist
\item
  We propose a \textbf{cofluctuation-based FC pretraining approach} that
  constructs retained and dropped views from framewise observations and
  couples consistency across retained views with retained-to-dropped latent
  prediction.
\item
  We develop \textbf{CoFlu-Brain}, an FC foundation model pretrained on
  multi-cohort fMRI data, and demonstrate strong frozen-transfer performance
  across demographic, developmental, and clinical downstream tasks.
\item
  We conduct a \textbf{systematic empirical analysis} across nine downstream
  tasks, examining the contributions of the learning objectives and the
  effects of data coverage, branch count, and temporal sampling, while
  assessing seed robustness for the primary and temporal-sampling comparisons.
\end{enumerate}

\begin{figure}[t]
\begin{minipage}[c]{0.62\linewidth}
\centering
\includegraphics[width=\linewidth]{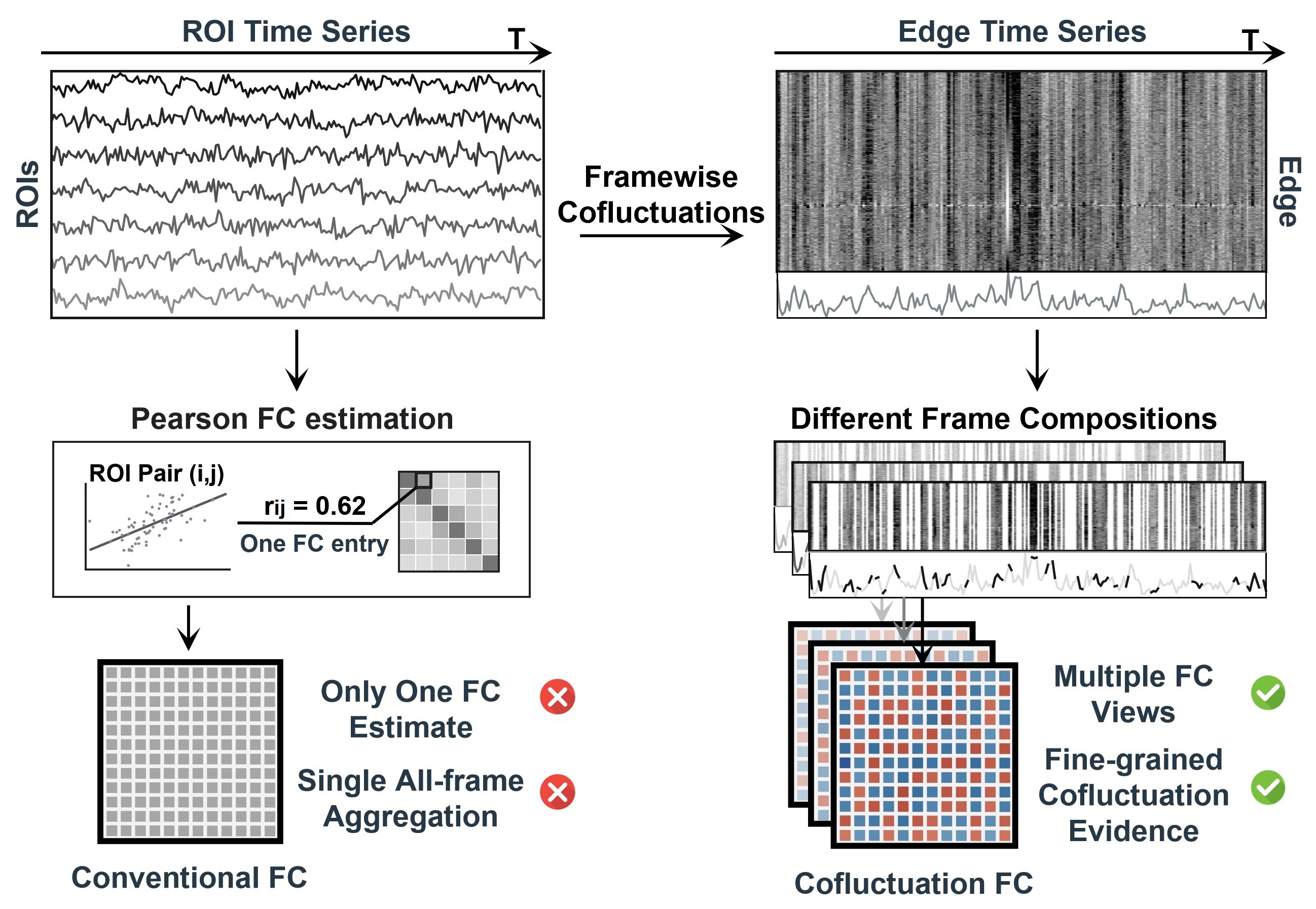}
\end{minipage}\hfill
\begin{minipage}[c]{0.34\linewidth}
\captionsetup{justification=raggedright,singlelinecheck=false}
\caption{\textbf{From a single FC estimate to cofluctuation based multi-view
learning.} Conventional FC aggregates all frames into a single connectivity
estimate, whereas CoFlu-Brain uses different compositions of framewise
cofluctuation observations to construct multiple FC views.}
\label{fig:motivation}
\end{minipage}
\end{figure}

\section{Related Work}\label{related-work}

fMRI representation models differ first in what they model. Time-series
methods reconstruct masked BOLD signals, predict masked latent targets, or
align representations across temporal views
\citep{brainlm,brainjepa,semantoks}. They preserve temporal dynamics, but
interregional connectivity remains an implicit property of the learned
activity representation.

Voxel-level models instead operate directly on 4D fMRI volumes
\citep{omnifmri,slimbrain,brainworld,flexibrain}. They retain fine-grained spatial
information without relying on a predefined atlas, but require
high-dimensional inputs and do not use connectivity as the primary modeling
unit.

FC models represent each scan as a connectivity matrix or graph, directly
encoding interregional coupling
\citep{kawahara2017brainnetcnn,li2021braingnn,brainnettransformer,brainmass,
braingfm,lcm}. Most treat the resulting Pearson-FC matrix as the object to
encode, reconstruct, augment, or align. This overlooks that FC is a
finite-sample estimate whose values depend on its supporting frames
\citep{noble2019review,esfahlani2020high,faskowitz2020edge}.
Dynamic-FC methods instead model temporal variation across windows or graph
sequences \citep{hutchison2013dynamic,kim2021stagin}, but address evolving
connectivity states rather than reproducibility across alternative estimates.

CoFlu-Brain retains the interpretability of FC modeling while extending its
learning unit from one scan-level estimate to multiple related estimates.
These views arise from complementary frame observations and preserve a direct
relationship to FC estimation, promoting representations stable across frame
compositions and predictive of held-out observations.

Self-supervised methods also differ in how they define their learning signal.
Masked autoencoding reconstructs omitted input content, whereas JEPA predicts
masked targets in latent space \citep{he2022masked,brainlm,brainmass,braingfm,
brainjepa,assran2023ijepa}. Joint-embedding methods such as BYOL and DINO
instead align representations across views, while VICReg and SIGReg regularize
the feature distribution against collapse
\citep{grill2020byol,caron2021dino,bardes2022vicreg,balestriero2025lejepa}.

Other directions include metadata-conditioned diffusion pretraining across
brain states \citep{braindit}, organizing pretraining domains and downstream
transfer through measured learning relations \citep{braintaskonomy}, and
jointly examining data size, model size, and training duration
\citep{fmriscaling}. These studies complement the design of FC-specific
training views explored here.

CoFlu-Brain integrates these principles through FC-specific view relations.
Consistency across retained estimates promotes stability under frame
resampling; prediction from retained to complementary dropped estimates
captures information predictive of held-out observations; and SIGReg
maintains representation diversity. Together, these components use the
finite-sample structure of FC to define which views should agree and which
complementary evidence should be predicted.

\section{Method}\label{method}

\subsection{Overview}\label{method-overview}

\begin{figure}[t]
\centering
\includegraphics[width=\linewidth,height=\textheight,keepaspectratio]{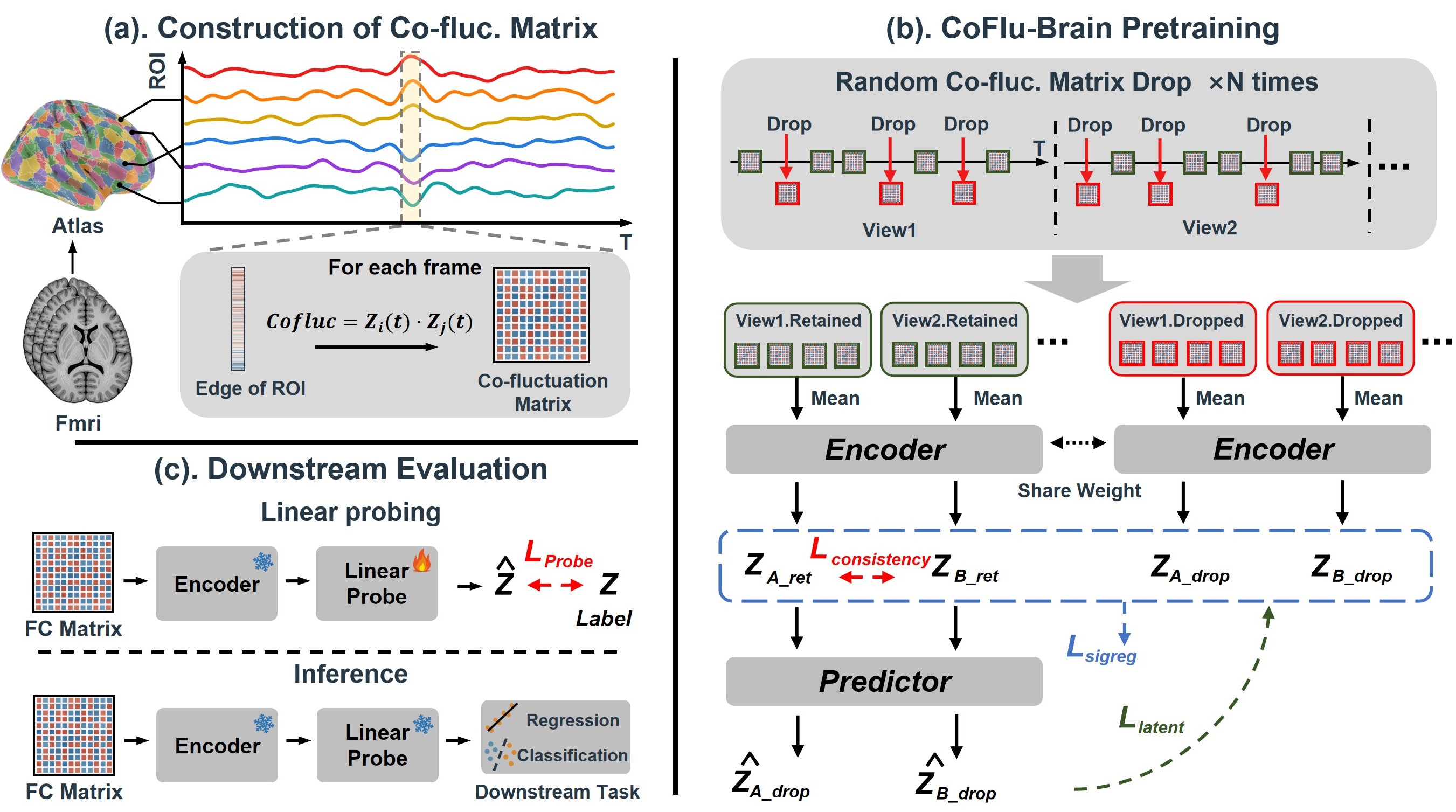}
\caption{\textbf{CoFlu-Brain pretraining and downstream evaluation.}
\textbf{(a)} Framewise cofluctuations from standardized regional BOLD signals.
\textbf{(b)} Paired retained--dropped FC views train a shared encoder through
consistency, latent prediction, and SIGReg.
\textbf{(c)} Frozen-encoder linear probing and downstream inference.}\label{fig:model}
\end{figure}

CoFlu-Brain constructs multiple retained--dropped FC pairs from each fMRI
segment. As illustrated in Figure~\ref{fig:model},
a shared encoder maps each estimate to a scan-level representation. Training
combines consistency across retained-frame representations, prediction of the
paired dropped-frame representation, and distributional regularization. After
pretraining, the projector and predictor are discarded, and the shared FC
encoder provides representations for downstream tasks.

\subsection{Cofluctuation-based retained and dropped views}
\label{cofluctuation-based-views}

Let \(X_b\in\mathbb{R}^{T\times N}\) denote the preprocessed fMRI segment of
sample \(b\). For notational simplicity, we describe the construction for a
fixed sample and suppress the sample index \(b\) below. For a selected frame
set \(A\), we independently standardize each regional signal within \(A\),
yielding \(z_t^A\) at frame \(t\). We define the framewise cofluctuation matrix
as

\[
E_t^A=z_t^A(z_t^A)^\top.
\]

CoFlu-Brain summarizes the cofluctuation observations associated with \(A\)
as an FC view:

\[
R^A_{ij}=\frac{1}{|A|-1}\sum_{t\in A}E^A_{t,ij},
\qquad i\neq j.
\]

This decomposition makes the frame support of each finite-sample FC
estimate explicit and enables the complementary view relations used below.

The resulting matrix is Fisher transformed on its off-diagonal entries before
being passed to the encoder. For each partition \(v\), we sample a retained
proportion \(r_v\). We draw the retained frame set \(S_v\) without replacement
and assign the remaining frames to the dropped set \(D_v\). The two sets do
not overlap within a pair, \(S_v\cap D_v=\varnothing\), while independently
sampled partitions may reuse frames across different pairs. We independently standardize the
retained and dropped sets and construct their cofluctuation-based FC views:

\[
C^{S_v}=\operatorname{FisherZ}\!\left(R^{S_v}\right),
\qquad
C^{D_v}=\operatorname{FisherZ}\!\left(R^{D_v}\right).
\]

The Fisher transformation is applied only to off-diagonal entries, and all
diagonal entries are set to zero. For a minibatch, we restore the sample index
\(b\), writing \(S_{vb}\), \(D_{vb}\), \(C_b^{S_{vb}}\), and
\(C_b^{D_{vb}}\).

The retained and dropped views summarize disjoint sets of framewise
cofluctuation observations from the same scan. Repeating this construction
for \(V\) independently sampled partitions exposes the model to different
cofluctuation compositions. The learning objective encourages connectivity
representations that remain consistent across retained views and predictive
of their paired dropped views.

\subsection{Model architecture}\label{shared-fc-representation}

The encoder represents each sampled frame set through its cofluctuation-based
FC estimate \(C\in\mathbb{R}^{N\times N}\). With diagonal entries set to zero,
we treat each row \(c_i\in\mathbb{R}^{N}\) as the signed connectivity profile
of ROI \(i\). A shared row-wise encoder
\(\phi:\mathbb{R}^{N}\rightarrow\mathbb{R}^{d}\) maps the \(N\) connectivity
profiles to \(N\) ROI tokens. A Vision Transformer (ViT) encoder \(F_\theta\)
\citep{dosovitskiy2021image} then contextualizes these tokens through
interactions among ROIs:

\[
H(C)=F_\theta\!\left(
[\phi(c_1),\ldots,\phi(c_N)]
\right)\in\mathbb{R}^{N\times d}.
\]

Mean pooling over the contextualized ROI tokens produces the scan-level FC
representation

\[
f_\theta(C)=\frac{1}{N}\sum_{i=1}^{N}H_i(C)
\in\mathbb{R}^{d}.
\]

For each of the \(V\) retained--dropped pairs and each sample \(b\) in a
minibatch of size \(B\), the same encoder and projector
\(g:\mathbb{R}^{d}\rightarrow\mathbb{R}^{d_p}\) are applied to both FC
estimates:

\[
p_{vb}=g\!\left(f_\theta(C_b^{S_{vb}})\right),\qquad
q_{vb}=g\!\left(f_\theta(C_b^{D_{vb}})\right),
\quad v=1,\ldots,V,\quad b=1,\ldots,B.
\]

A predictor \(h:\mathbb{R}^{d_p}\rightarrow\mathbb{R}^{d_p}\) maps each
retained-frame projection to \(\hat q_{vb}=h(p_{vb})\), which is trained to
predict the corresponding dropped-frame projection \(q_{vb}\). The retained
and dropped branches share all encoder and projector parameters, and both
receive gradients during pretraining. After pretraining, the projector and
predictor are discarded; only the encoder representation \(f_\theta(C)\) is
used for frozen downstream probing. For downstream evaluation, \(C\) was
computed from all 200 frames of each input segment. Architectural dimensions
and optimization settings are reported in
Appendix~\ref{appendix-training-and-probing}.

\subsection{Learning across FC estimates}
\label{learning-across-fc-estimates}

\paragraph{Reproducibility across retained estimates.}
Retained-frame FC matrices from the same scan differ numerically because each
is supported by a different collection of cofluctuations. We encourage their
representations to preserve the structure that can be inferred reproducibly
across estimates:

\[
\bar p_b=\frac{1}{V}\sum_{v=1}^{V}p_{vb},\qquad
\mathcal{L}_{\mathrm{con}}=
\frac{1}{VBd_p}\sum_{v=1}^{V}\sum_{b=1}^{B}
\left\|p_{vb}-\bar p_b\right\|_2^2 .
\]

Here, \(V\), \(B\), and \(d_p\) denote the number of temporal partitions, the
minibatch size, and the projection dimension, respectively.

\paragraph{Prediction of dropped-frame evidence.}
Consistency captures structure shared across retained estimates. To also learn
across paired non-overlapping frame subsets, a predictor \(h\) infers the
representation supported by the dropped frames:

\[
\mathcal{L}_{\mathrm{pred}}=
\frac{1}{VB}\sum_{v=1}^{V}\sum_{b=1}^{B}
\left[1-\cos\!\left(h(p_{vb}),q_{vb}\right)\right].
\]

Both branches share parameters and receive gradients without requiring a
stop-gradient target or momentum encoder. This term trains each retained FC
representation to predict its paired dropped-frame representation.

\paragraph{Non-collapse regularization.}
Consistency and prediction specify relationships among paired FC estimates.
We additionally apply SIGReg to the collection of retained and dropped
projections following LeJEPA \citep{balestriero2025lejepa}, regularizing their
distribution against representational collapse. The complete objective is

\[
\mathcal{L}=
\mathcal{L}_{\mathrm{pred}}
+\lambda_{\mathrm{con}}\gamma(e)\mathcal{L}_{\mathrm{con}}
+\lambda_{\mathrm{reg}}\mathcal{L}_{\mathrm{SIGReg}},
\]

where \(\gamma(e)\) controls the contribution of consistency through an initial
full-weight phase followed by linear decay:
\[
\gamma(e)=
\begin{cases}
1, & e\leq E_w,\\[2pt]
\max\!\left(1-\dfrac{e-E_w}{E_d},0\right), & e>E_w.
\end{cases}
\]
Here, \(e\) is the training epoch, \(E_w\) is the duration of the initial
full-weight phase, and \(E_d\) is the subsequent decay duration. In the main
configuration, \(E_w=1\) and \(E_d=24\), so the consistency weight reaches
zero at epoch 25. The schedule emphasizes alignment early in training and
shifts the later objective toward dropped-frame prediction. The three loss
terms respectively promote latent prediction, reproducibility across finite
FC estimates, and a non-degenerate representation distribution.

\section{Results}\label{results}

\setlength{\intextsep}{4pt}
\setlength{\textfloatsep}{8pt}
\setlength{\floatsep}{6pt}

\subsection{Experimental setup}\label{experimental-setup}
\label{pretraining-data-and-training-contract}
\label{pretraining-protocol}
\label{downstream-evaluation}

CoFlu-Brain was pretrained on 13,444 subjects from eight dataset groups:
CHCP, HCP, ABCD, ABCC, HBN, PNC, MDD, and AOMIC PIOP1/2
\citep{ge2023chcp,vanessen2013hcp,casey2018abcd,feczko2021abcc,
alexander2017hbn,satterthwaite2014pnc,yan2019restmdd,snoek2021aomic}.
All inputs were represented as Schaefer-400 cortical time series
\citep{schaefer2018local}, and 200-frame segments were used for pretraining.
Downstream evaluation comprised two in-cohort tasks from HCP and ABCD, whose
unlabeled scans contributed to pretraining, and seven out-of-cohort tasks from
SALD, PPMI, ADNI, ABIDE, and ADHD200, which were excluded from pretraining
\citep{vanessen2013hcp,casey2018abcd,wei2018sald,marek2011ppmi,
weiner2010adni,dimartino2014abide,adhd200consortium2012}.
No downstream labels were used during pretraining.

Frozen representations were evaluated with logistic regression for
classification and ridge regression for age prediction. For age prediction,
nMAE was computed after standardizing the target using the training-fold mean
and population standard deviation. Cross-dataset ranks used weighted F1 for
classification and nMAE for age prediction. Endpoint ranks were averaged
within each cohort, and the dispersion of method mean ranks was evaluated
using two million method-label permutations within cohort blocks. Full details
are provided in Appendix~\ref{appendix-experimental-details}.

\subsection{Main Results}\label{existing-model-comparison}

We evaluated CoFlu-Brain against four released fMRI representation models:
BrainGFM, BrainMASS, BrainLM, and the Large Connectome Model
\citep{braingfm,brainmass,brainlm,lcm}.

\begin{table}[!ht]
\centering
\caption{\textbf{Linear probing of fMRI models for classification and age
prediction.} Values are mean \(\pm\) sample SD
across five participant-grouped folds. Classification tasks report accuracy
and weighted F1 (\(\uparrow\)); age-prediction tasks report nMAE
(\(\downarrow\)) and Pearson's \(r\) (\(\uparrow\)). Red boldface marks the
best mean for each metric. CoFlu-Brain averages three pretrained checkpoints
per fold; baselines use one released checkpoint. Overall rank test:
cohort-blocked permutation \(p<0.001\).}
\label{tab:existing-model-transfer}
\definecolor{TableGroup}{RGB}{242,242,242}
\definecolor{CoFluTint}{RGB}{239,239,239}
\scriptsize
\setlength{\tabcolsep}{4.0pt}
\renewcommand{\arraystretch}{1.12}
\newcommand{\taskdivider}{%
  \raisebox{-2.3pt}[0pt][0pt]{%
  \ensuremath{\left.\rule{0pt}{7.5pt}\right|}}}
\resizebox{\linewidth}{!}{%
\begin{tabular}{@{}c@{\hspace{4pt}}c@{\hspace{4pt}}c>{\columncolor{CoFluTint}}ccccc@{}}
\toprule
\multicolumn{3}{@{}c}{%
\setlength{\unitlength}{1pt}%
\begin{picture}(130,26)
\put(0,26){\line(5,-1){130}}
\put(94,20){\makebox(0,0)[c]{\footnotesize\bfseries Model}}
\put(36,6){\makebox(0,0)[c]{\footnotesize\bfseries Dataset}}
\end{picture}} &
\raisebox{10pt}{\footnotesize\bfseries CoFlu-Brain} &
\raisebox{10pt}{\footnotesize\bfseries BrainGFM} &
\raisebox{10pt}{\footnotesize\bfseries BrainMASS} &
\raisebox{10pt}{\footnotesize\bfseries BrainLM} &
\raisebox{10pt}{\footnotesize\bfseries LCM} \\
\hline
\noalign{\vskip 1.5pt}
\multirow{2}{*}{\bfseries HCP / Sex}
 & \taskdivider & Accuracy \(\uparrow\) & \best{0.7749 \(\pm\) 0.0111} & 0.6428 \(\pm\) 0.0123 & 0.7207 \(\pm\) 0.0053 & 0.6258 \(\pm\) 0.0138 & 0.7292 \(\pm\) 0.0163 \\
 & & Weighted F1 \(\uparrow\) & \best{0.7745 \(\pm\) 0.0103} & 0.6432 \(\pm\) 0.0119 & 0.7203 \(\pm\) 0.0048 & 0.6258 \(\pm\) 0.0129 & 0.7284 \(\pm\) 0.0157 \\
\multirow{2}{*}{\bfseries ABCD / Sex}
 & \taskdivider & Accuracy \(\uparrow\) & 0.5953 \(\pm\) 0.0195 & 0.5336 \(\pm\) 0.0482 & \best{0.6216 \(\pm\) 0.0516} & 0.4965 \(\pm\) 0.0352 & 0.5436 \(\pm\) 0.0308 \\
 & & Weighted F1 \(\uparrow\) & 0.5961 \(\pm\) 0.0190 & 0.5317 \(\pm\) 0.0528 & \best{0.6215 \(\pm\) 0.0515} & 0.4967 \(\pm\) 0.0350 & 0.5420 \(\pm\) 0.0331 \\
\multirow{2}{*}{\bfseries PPMI / Group}
 & \taskdivider & Accuracy \(\uparrow\) & \best{0.5918 \(\pm\) 0.0063} & 0.5871 \(\pm\) 0.0335 & 0.5813 \(\pm\) 0.0280 & 0.5269 \(\pm\) 0.0192 & 0.5501 \(\pm\) 0.0262 \\
 & & Weighted F1 \(\uparrow\) & \best{0.5775 \(\pm\) 0.0173} & 0.5502 \(\pm\) 0.0468 & 0.5751 \(\pm\) 0.0301 & 0.5237 \(\pm\) 0.0300 & 0.5354 \(\pm\) 0.0373 \\
\multirow{2}{*}{\bfseries ADNI / AD}
 & \taskdivider & Accuracy \(\uparrow\) & 0.6841 \(\pm\) 0.0306 & \best{0.6870 \(\pm\) 0.0194} & 0.5826 \(\pm\) 0.0418 & 0.6217 \(\pm\) 0.1049 & 0.5957 \(\pm\) 0.1136 \\
 & & Weighted F1 \(\uparrow\) & \best{0.6835 \(\pm\) 0.0307} & 0.6832 \(\pm\) 0.0236 & 0.5811 \(\pm\) 0.0393 & 0.6183 \(\pm\) 0.1078 & 0.5934 \(\pm\) 0.1179 \\
\multirow{2}{*}{\bfseries ADNI / MCI}
 & \taskdivider & Accuracy \(\uparrow\) & \best{0.5570 \(\pm\) 0.0415} & 0.5515 \(\pm\) 0.0351 & 0.5070 \(\pm\) 0.0339 & 0.5412 \(\pm\) 0.0661 & 0.5513 \(\pm\) 0.0233 \\
 & & Weighted F1 \(\uparrow\) & \best{0.5538 \(\pm\) 0.0441} & 0.5480 \(\pm\) 0.0367 & 0.5045 \(\pm\) 0.0346 & 0.5367 \(\pm\) 0.0673 & 0.5506 \(\pm\) 0.0232 \\
\multirow{2}{*}{\bfseries ABIDE / ASD}
 & \taskdivider & Accuracy \(\uparrow\) & \best{0.5602 \(\pm\) 0.0474} & 0.4911 \(\pm\) 0.0734 & 0.5487 \(\pm\) 0.0520 & 0.5087 \(\pm\) 0.1294 & 0.4971 \(\pm\) 0.0618 \\
 & & Weighted F1 \(\uparrow\) & \best{0.5567 \(\pm\) 0.0476} & 0.4902 \(\pm\) 0.0737 & 0.5472 \(\pm\) 0.0527 & 0.5057 \(\pm\) 0.1333 & 0.4913 \(\pm\) 0.0630 \\
\multirow{2}{*}{\bfseries ADHD200 / Diagnosis}
 & \taskdivider & Accuracy \(\uparrow\) & \best{0.5899 \(\pm\) 0.0155} & 0.5297 \(\pm\) 0.0313 & 0.5559 \(\pm\) 0.0257 & 0.5241 \(\pm\) 0.0105 & 0.5797 \(\pm\) 0.0397 \\
 & & Weighted F1 \(\uparrow\) & \best{0.5891 \(\pm\) 0.0151} & 0.5286 \(\pm\) 0.0285 & 0.5549 \(\pm\) 0.0254 & 0.5235 \(\pm\) 0.0089 & 0.5806 \(\pm\) 0.0381 \\
\multirow{2}{*}{\bfseries SALD / Age}
 & \taskdivider & nMAE \(\downarrow\) & \best{0.6571 \(\pm\) 0.0431} & 0.8548 \(\pm\) 0.0605 & 0.9064 \(\pm\) 0.0877 & 0.9824 \(\pm\) 0.0473 & 0.8759 \(\pm\) 0.0641 \\
 & & Pearson \(r\) \(\uparrow\) & \best{0.6422 \(\pm\) 0.0418} & 0.4262 \(\pm\) 0.0600 & 0.4415 \(\pm\) 0.1125 & 0.3180 \(\pm\) 0.1073 & 0.4335 \(\pm\) 0.0793 \\
\multirow{2}{*}{\bfseries ADHD200 / Age}
 & \taskdivider & nMAE \(\downarrow\) & \best{0.6557 \(\pm\) 0.0370} & 0.8406 \(\pm\) 0.0797 & 0.9328 \(\pm\) 0.0377 & 1.1863 \(\pm\) 0.0455 & 0.9135 \(\pm\) 0.0545 \\
 & & Pearson \(r\) \(\uparrow\) & \best{0.5967 \(\pm\) 0.0249} & 0.2696 \(\pm\) 0.0593 & 0.3525 \(\pm\) 0.0526 & 0.2374 \(\pm\) 0.0783 & 0.4433 \(\pm\) 0.0476 \\
\midrule
\multicolumn{3}{@{}l}{\bfseries Cohort-averaged rank} &
\best{1.14} & 3.36 & 2.93 & 4.50 & 3.07 \\
\noalign{\vskip -1.5pt}
\bottomrule
\end{tabular}
}
\end{table}

CoFlu-Brain led eight of nine tasks and achieved the best rank across the
seven evaluation cohorts (1.14; Table~\ref{tab:existing-model-transfer}).

\subsection{Functional-network analysis on ADHD200}\label{network-restricted-interpretation}

\begin{figure}[!t]
\centering
\includegraphics[width=\linewidth,height=0.46\textheight,keepaspectratio]{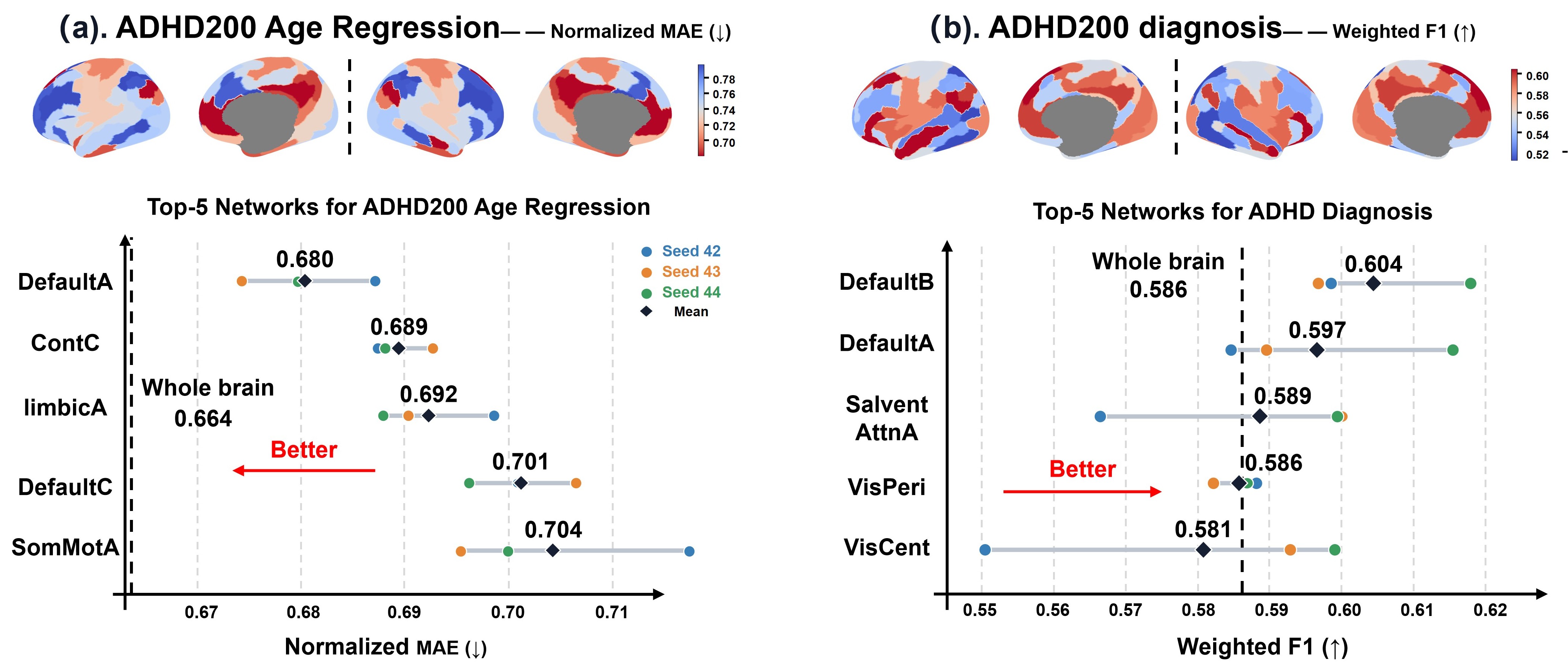}
\caption{\textbf{Network-restricted probing on ADHD200.}
\textbf{(a)} Age nMAE (\(\downarrow\)).
\textbf{(b)} Diagnosis weighted F1 (\(\uparrow\)).
Cortical maps summarize mean performance across all 17 networks, and the
corresponding point-range plots detail the five best-performing networks for
each task. Points indicate individual pretraining seeds, diamonds indicate
seed means, horizontal segments indicate seed ranges, and dashed vertical
lines indicate whole-brain performance.}\label{fig:adhd-network-probing}
\end{figure}

We examined how predictive information for age and ADHD diagnosis was
distributed across the Yeo-17 functional systems \citep{yeo2011organization}.
For each network, we
retained its subject-specific within-network FC, replaced the remaining
entries with their probe-training means, and evaluated the resulting
representation using the frozen encoder and linear probe
(Figure~\ref{fig:adhd-network-probing}).

Network rankings were consistent across the three pretrained encoders, with
mean pairwise rank correlations of \(\rho=0.906\) for age and \(\rho=0.857\)
for diagnosis. The prominence of default-network components in both tasks is
consistent with their progressive integration during brain maturation
\citep{dosenbach2010maturity,fair2008default}. Their joint
contribution with the salience/ventral-attention system to diagnosis further
reflects the distributed network organization associated with ADHD
\citep{sripada2014lag,sripada2014disrupted,norman2023default}
(see Appendix~\ref{appendix-complete-network-probing} for complete results).

\subsection{Ablation and sensitivity analyses}\label{ablation-studies}

We evaluated temporal sampling
(Appendix~\ref{appendix-temporal-sampling-details}) and pretraining objectives
(Appendix~\ref{appendix-a.-objective-component-ablation-on-representative-tasks})
through ablations, and examined sensitivity to data coverage
(Appendix~\ref{appendix-b.-data-coverage-scaling-details}), temporal branch
count (Appendix~\ref{appendix-c.-branch-count-scaling-details}), and
consistency-weight scheduling
(Appendix~\ref{appendix-d.-loss-weight-and-schedule-sensitivity}).
For compact presentation, all four tabular analyses use the same task subset;
the appendices report all nine tasks.

\subsubsection{Random versus ordered frame partitioning}
\label{temporal-sampling-ablation}

To compare temporal sampling strategies, we evaluated random and ordered partitions at
matched retained-frame ratios of \(0.4\)--\(0.7\). Random partitioning sampled
retained frames throughout the sequence and used the complement as the dropped
view; ordered partitioning used the first \(k\) and remaining \(T-k\) frames.
Both used identical standardization, FC construction, architecture, and
objective, differing only in frame indices. This comparison assesses the
empirical effect of the two frame-partitioning strategies on downstream
transfer.

\begin{figure}[!t]
\centering
\includegraphics[width=\linewidth]{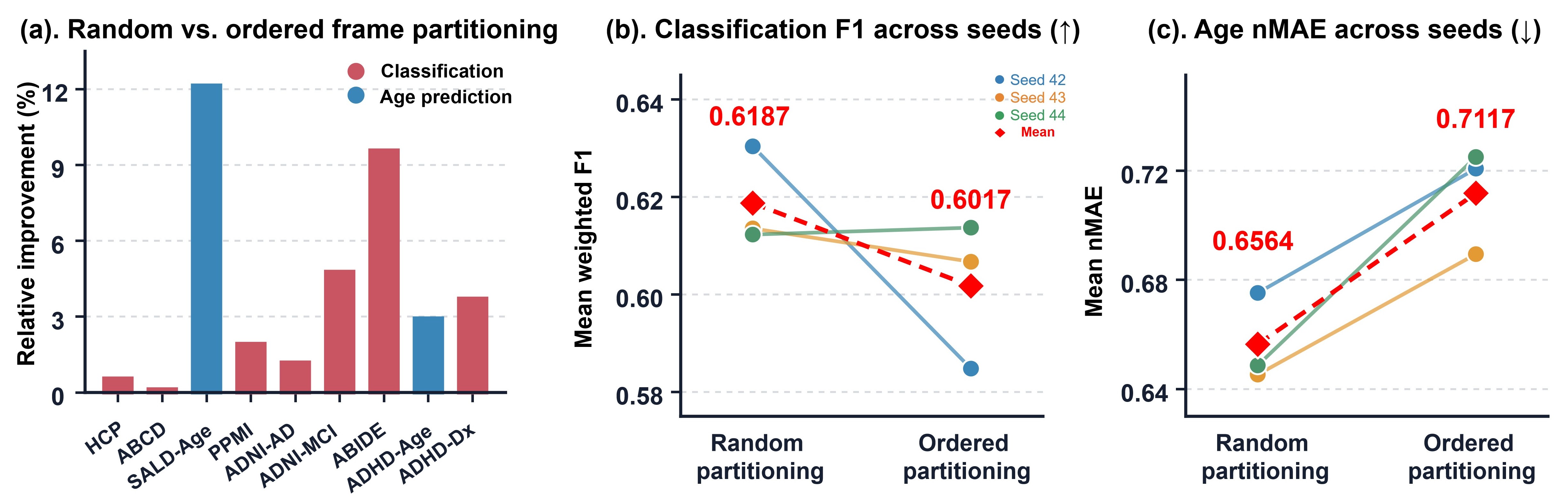}
\caption{\textbf{Random versus ordered frame partitioning.}
\textbf{(a)} Relative weighted-F1 gain for classification and relative nMAE
reduction for age prediction; positive values favor random partitioning.
\textbf{(b)} Mean weighted F1 across seven classification tasks.
\textbf{(c)} Mean nMAE across two age-prediction tasks.
Lines connect paired pretraining seeds; red diamonds denote seed means.}
\label{fig:temporal-sampling-main}
\end{figure}

On seed-averaged performance, random partitioning outperformed ordered
partitioning across all seven evaluation cohorts (exact sign test,
\(p=0.016\); Figure~\ref{fig:temporal-sampling-main}).

\subsubsection{Pretraining objectives}
\label{objective-component-ablation-and-representation-diagnostics}

We next assessed the contribution of the three objective components by
removing prediction, consistency, or SIGReg while preserving the remaining
pretraining configuration. Table~\ref{tab:objective-ablation} presents four
downstream tasks, while
Appendix~\ref{appendix-a.-objective-component-ablation-on-representative-tasks}
reports the complete nine-task results and representation diagnostics.

\begin{table}[!htbp]
\centering
\caption{\textbf{Downstream performance under objective-component ablation.}
Each objective variant was pretrained independently with seed 42 under matched
configurations and is reported as mean \(\pm\) SD across five participant-grouped folds.
Metrics are weighted F1 (\(\uparrow\)) and age nMAE (\(\downarrow\)).
Red boldface marks the best mean for each task. Overall rank test:
cohort-blocked permutation \(p=0.0014\).}
\label{tab:objective-ablation}
\definecolor{FullTint}{RGB}{239,239,239}
\scriptsize
\setlength{\tabcolsep}{4pt}
\renewcommand{\arraystretch}{1.15}
\resizebox{\linewidth}{!}{%
\begin{tabular}{@{}>{\centering\arraybackslash}m{78pt}cccc@{}}
\toprule
\multirow{2}{*}{%
\setlength{\unitlength}{1pt}%
\begin{picture}(78,16)
\put(0,16){\line(5,-1){78}}
\put(58,12){\makebox(0,0)[c]{\scriptsize\bfseries Dataset}}
\put(19,2){\makebox(0,0)[c]{\scriptsize\bfseries Objective}}
\end{picture}} &
\bfseries HCP / Sex &
\bfseries ADNI / AD &
\bfseries ADHD200 / Diagnosis &
\bfseries ADHD200 / Age \\
\cmidrule(lr){2-2}\cmidrule(lr){3-3}\cmidrule(lr){4-4}\cmidrule(l){5-5}
\noalign{\vskip-2pt}
&
Weighted F1 \(\uparrow\) &
Weighted F1 \(\uparrow\) &
Weighted F1 \(\uparrow\) &
nMAE \(\downarrow\) \\[1pt]
\midrule
\rowcolor{FullTint}
\bfseries Full &
\best{0.7874 \(\pm\) 0.0115} &
\best{0.7219 \(\pm\) 0.0657} &
\best{0.5872 \(\pm\) 0.0215} &
\best{0.6523 \(\pm\) 0.0315} \\
\bfseries No prediction &
0.6565 \(\pm\) 0.0187 &
0.5023 \(\pm\) 0.0850 &
0.5348 \(\pm\) 0.0309 &
0.7416 \(\pm\) 0.0567 \\
\bfseries No consistency &
0.7789 \(\pm\) 0.0110 &
0.6866 \(\pm\) 0.1175 &
0.5725 \(\pm\) 0.0384 &
0.6611 \(\pm\) 0.0539 \\
\bfseries No SIGReg &
0.6942 \(\pm\) 0.0153 &
0.6065 \(\pm\) 0.0667 &
0.5643 \(\pm\) 0.0288 &
0.7132 \(\pm\) 0.0482 \\
\midrule
\multicolumn{5}{@{}c@{}}{\bfseries Cohort-averaged rank:
Full 1.50;\quad No prediction 3.57;\quad No consistency 1.79;\quad No SIGReg 3.14} \\
\bottomrule
\end{tabular}
}
\end{table}

The complete objective ranked best overall (1.50;
Table~\ref{tab:objective-ablation}). Removing prediction or SIGReg generally
reduced transfer, while consistency produced task-dependent gains.

\subsubsection{Pretraining data coverage}\label{data-coverage-comparison}

To examine how the breadth of pretraining data affected transfer, we increased
the pretraining manifest from 12.5\% to 100\% while keeping the architecture
and number of optimization updates fixed. Table~\ref{tab:data-coverage-main}
shows four downstream tasks, and
Appendix~\ref{appendix-b.-data-coverage-scaling-details} reports all nine.

\begin{table}[!htbp]
\centering
\caption{\textbf{Pretraining data coverage and downstream performance.}
Values are mean \(\pm\) SD across probe folds.
Metrics are weighted F1 (\(\uparrow\)) and age nMAE (\(\downarrow\)).
Red boldface marks the best mean for each task.}
\label{tab:data-coverage-main}
\definecolor{MainTint}{RGB}{239,239,239}
\scriptsize
\setlength{\tabcolsep}{4pt}
\renewcommand{\arraystretch}{1.15}
\resizebox{\linewidth}{!}{%
\begin{tabular}{@{}>{\centering\arraybackslash}m{78pt}cccc@{}}
\toprule
\multirow{2}{*}{%
\setlength{\unitlength}{1pt}%
\begin{picture}(78,16)
\put(0,16){\line(5,-1){78}}
\put(58,12){\makebox(0,0)[c]{\scriptsize\bfseries Dataset}}
\put(19,2){\makebox(0,0)[c]{\scriptsize\bfseries Coverage}}
\end{picture}} &
\bfseries HCP / Sex &
\bfseries ADNI / AD &
\bfseries ADHD200 / Diagnosis &
\bfseries ADHD200 / Age \\
\cmidrule(lr){2-2}\cmidrule(lr){3-3}\cmidrule(lr){4-4}\cmidrule(l){5-5}
\noalign{\vskip-2pt}
&
Weighted F1 \(\uparrow\) &
Weighted F1 \(\uparrow\) &
Weighted F1 \(\uparrow\) &
nMAE \(\downarrow\) \\[1pt]
\midrule
\bfseries 12.5\% &
0.7265 \(\pm\) 0.0126 &
0.6627 \(\pm\) 0.0137 &
0.5678 \(\pm\) 0.0134 &
0.7553 \(\pm\) 0.0318 \\
\bfseries 25\% &
0.7376 \(\pm\) 0.0117 &
\best{0.7280 \(\pm\) 0.0348} &
0.5609 \(\pm\) 0.0097 &
0.7453 \(\pm\) 0.0341 \\
\bfseries 50\% &
0.7326 \(\pm\) 0.0240 &
0.6786 \(\pm\) 0.0301 &
0.5314 \(\pm\) 0.0333 &
0.7152 \(\pm\) 0.0174 \\
\rowcolor{MainTint}
\bfseries 100\% &
\best{0.7737 \(\pm\) 0.0092} &
0.7235 \(\pm\) 0.0336 &
\best{0.5756 \(\pm\) 0.0298} &
\best{0.6785 \(\pm\) 0.0224} \\
\bottomrule
\end{tabular}
}
\end{table}

Using the complete manifest improved all four tasks relative to 12.5\%
coverage and produced the best result on three of them
(Table~\ref{tab:data-coverage-main}).

\subsubsection{Number of temporal branches}\label{branch-count-comparison}

Each temporal branch provides a distinct retained-frame FC estimate during
pretraining. We varied the number of branches from two to eight to determine
how many retained--dropped estimate pairs were most effective, while retaining the
same update budget and normalizing the consistency objective across branch
counts. Table~\ref{tab:branch-count-main} shows four downstream tasks;
Appendix~\ref{appendix-c.-branch-count-scaling-details} reports all nine
together with training-time details.

\begin{table}[!htbp]
\centering
\caption{\textbf{Downstream performance by temporal branch count.}
Values are mean \(\pm\) SD across probe folds.
Metrics are weighted F1 (\(\uparrow\)) and age nMAE (\(\downarrow\)).
Red boldface marks the best mean for each task.}
\label{tab:branch-count-main}
\definecolor{MainTint}{RGB}{239,239,239}
\scriptsize
\setlength{\tabcolsep}{4pt}
\renewcommand{\arraystretch}{1.15}
\resizebox{\linewidth}{!}{%
\begin{tabular}{@{}>{\centering\arraybackslash}m{78pt}cccc@{}}
\toprule
\multirow{2}{*}{%
\setlength{\unitlength}{1pt}%
\begin{picture}(78,16)
\put(0,16){\line(5,-1){78}}
\put(58,12){\makebox(0,0)[c]{\scriptsize\bfseries Dataset}}
\put(19,2){\makebox(0,0)[c]{\scriptsize\bfseries Branches}}
\end{picture}} &
\bfseries HCP / Sex &
\bfseries ADNI / AD &
\bfseries ADHD200 / Diagnosis &
\bfseries ADHD200 / Age \\
\cmidrule(lr){2-2}\cmidrule(lr){3-3}\cmidrule(lr){4-4}\cmidrule(l){5-5}
\noalign{\vskip-2pt}
&
Weighted F1 \(\uparrow\) &
Weighted F1 \(\uparrow\) &
Weighted F1 \(\uparrow\) &
nMAE \(\downarrow\) \\[1pt]
\midrule
\bfseries B2 &
0.7557 \(\pm\) 0.0205 &
0.6518 \(\pm\) 0.0536 &
\best{0.5762 \(\pm\) 0.0387} &
0.7088 \(\pm\) 0.0091 \\
\rowcolor{MainTint}
\bfseries B4 &
\best{0.7737 \(\pm\) 0.0092} &
\best{0.7235 \(\pm\) 0.0336} &
0.5756 \(\pm\) 0.0298 &
\best{0.6785 \(\pm\) 0.0224} \\
\bfseries B6 &
0.7601 \(\pm\) 0.0296 &
0.6429 \(\pm\) 0.0818 &
0.5637 \(\pm\) 0.0183 &
0.6872 \(\pm\) 0.0195 \\
\bfseries B8 &
0.7643 \(\pm\) 0.0122 &
0.6686 \(\pm\) 0.0470 &
0.5713 \(\pm\) 0.0269 &
0.6894 \(\pm\) 0.0338 \\
\bottomrule
\end{tabular}
}
\end{table}

Four branches achieved the best performance on three of the four tasks,
whereas increasing from four to eight branches raised the pretraining cost
from 10.54 to 26.72 GPU hours without consistent performance gains
(Table~\ref{tab:branch-count-main}).

\subsubsection{Consistency-weight schedule}
\label{consistency-decay-schedule-ablation}

Consistency aligns retained-frame compositions, while prediction captures
relationships between retained- and dropped-frame FC estimates. We
hypothesized that emphasizing consistency early and prediction later could
balance these signals. We therefore compared Decay-25, Decay-50, and Constant
consistency schedules. Table~\ref{tab:decay-main} presents the three
schedules, while
Appendix~\ref{appendix-d.-loss-weight-and-schedule-sensitivity} reports their
performance on all nine downstream tasks.

\begin{table}[!htbp]
\centering
\caption{\textbf{Downstream performance under consistency-weight schedules.}
Values are mean \(\pm\) SD across probe folds.
Metrics are weighted F1 (\(\uparrow\)) and age nMAE (\(\downarrow\)).
Red boldface marks the best mean for each task.}
\label{tab:decay-main}
\definecolor{MainTint}{RGB}{239,239,239}
\scriptsize
\setlength{\tabcolsep}{4pt}
\renewcommand{\arraystretch}{1.15}
\resizebox{\linewidth}{!}{%
\begin{tabular}{@{}>{\centering\arraybackslash}m{78pt}cccc@{}}
\toprule
\multirow{2}{*}{%
\setlength{\unitlength}{1pt}%
\begin{picture}(78,16)
\put(0,16){\line(5,-1){78}}
\put(58,12){\makebox(0,0)[c]{\scriptsize\bfseries Dataset}}
\put(19,2){\makebox(0,0)[c]{\scriptsize\bfseries Schedule}}
\end{picture}} &
\bfseries HCP / Sex &
\bfseries ADNI / AD &
\bfseries ADHD200 / Diagnosis &
\bfseries ADHD200 / Age \\
\cmidrule(lr){2-2}\cmidrule(lr){3-3}\cmidrule(lr){4-4}\cmidrule(l){5-5}
\noalign{\vskip-2pt}
&
Weighted F1 \(\uparrow\) &
Weighted F1 \(\uparrow\) &
Weighted F1 \(\uparrow\) &
nMAE \(\downarrow\) \\[1pt]
\midrule
\rowcolor{MainTint}
\bfseries Decay-25 &
\best{0.7737 \(\pm\) 0.0092} &
\best{0.7235 \(\pm\) 0.0336} &
\best{0.5756 \(\pm\) 0.0298} &
0.6785 \(\pm\) 0.0224 \\
\bfseries Decay-50 &
0.7588 \(\pm\) 0.0204 &
0.6736 \(\pm\) 0.0236 &
0.5433 \(\pm\) 0.0282 &
0.6839 \(\pm\) 0.0239 \\
\bfseries Constant &
0.7687 \(\pm\) 0.0164 &
0.6627 \(\pm\) 0.0350 &
0.5688 \(\pm\) 0.0300 &
\best{0.6568 \(\pm\) 0.0270} \\
\bottomrule
\end{tabular}
}
\end{table}

Decay-25 performed best on all three classification tasks, whereas the
constant schedule performed best for age prediction
(Table~\ref{tab:decay-main}).

\section{Discussion}\label{discussion}

CoFlu-Brain shows that relationships among FC estimates supported by different
frame compositions provide useful self-supervision. Its representations
transferred across demographic, developmental, and clinical tasks, while
random partitioning improved aggregate transfer. The complete objective
combines retained-view consistency, dropped-view prediction, and
distributional regularization as complementary signals.

\paragraph{Limitations and future directions.}
CoFlu-Brain uses the framewise cofluctuation decomposition of Pearson
connectivity to construct multiple FC estimates with explicitly controlled
frame support. Because each view aggregates the selected framewise products
into a subset-specific FC matrix, the current model does not preserve the
timing, order, or amplitude of individual cofluctuation events. Its scope is
instead to learn from structured relationships among finite-sample FC
estimates: reproducibility across retained-frame compositions and prediction
of complementary held-out estimates. Future work could combine this
estimate-level objective with order-aware or edge-time-series encoders to
capture transient dynamics. Biological validation, causal analysis, and
integration with voxel-level features remain important directions beyond the
predictive evaluations considered here.

\section*{AI use statement}

Generative AI tools were used to assist manuscript organization and language
editing, table formatting, and consistency auditing between reported results
and experiment records. They were not used to generate experimental data. The
authors reviewed all AI-assisted text, independently verified the reported
numerical values against the underlying result files, and take responsibility
for the final content of the submission.

\section*{Ethics statement}

This work performs secondary analysis of de-identified neuroimaging data from
established research cohorts and does not recruit new participants or attempt
participant re-identification. Data access and use are governed by the terms of
the respective source datasets. Cohort composition, preprocessing provenance,
and pretraining-data exposure are documented to support appropriately scoped
interpretation of the transfer results.

\section*{Reproducibility statement}

Section~\ref{method} specifies the model and objective, and
Section~\ref{experimental-setup} defines the pretraining and downstream
evaluation protocol. Supplementary
Appendix~\ref{appendix-experimental-details} documents the datasets,
preprocessing, baselines, architecture, and training and probing
configurations. Appendix~\ref{appendix-expanded-main-results} provides
seed-wise and complete network-level results, while
Appendix~\ref{appendix-ablation-and-sensitivity} reports the full ablation and
sensitivity analyses. The implementation, configuration files, verification
scripts, and pretrained checkpoints are available in the anonymous
repository:
\url{https://anonymous.4open.science/r/review-artifact-2027/}.

\renewcommand\refname{References}
\bibliography{references.bib}

\clearpage
\appendix
\section{Experimental Details}\label{appendix-experimental-details}

\subsection{Datasets and downstream tasks}
\label{appendix-datasets-and-tasks}

\paragraph{Pretraining datasets.}
The pretraining corpus contained 13,444 usable subjects from eight dataset
groups. Only fMRI inputs, without downstream labels, were used during
pretraining.

\paragraph{Preprocessing.}
We started from cohort-specific preprocessed derivatives: the HCP minimal
preprocessing pipeline for HCP and CHCP, ABCD-HCP derivatives for ABCD and
ABCC, fMRIPrep derivatives for AOMIC PIOP1/2, and available MNI-space
preprocessed derivatives for HBN, PNC, and the multisite MDD data. Thus,
low-level operations, including motion correction and any nuisance-signal
handling, were inherited from the corresponding source pipeline rather than
re-estimated with a single cross-cohort pipeline. We then applied a common
input harmonization procedure: all fMRI volumes were represented at 2-mm
isotropic resolution, and only datasets with a source TR different from
0.72\,s were temporally interpolated to 0.72\,s using cubic B-spline
interpolation. Regional signals were extracted using the Schaefer-400 atlas,
and each regional time series was Z-standardized along the temporal dimension.
Files with exactly 200 frames were used directly; during pretraining,
contiguous 200-frame windows were sampled from longer series, whereas
sequences shorter than 200 frames were excluded without zero-padding.

\paragraph{Chinese Human Connectome Project (CHCP).}
CHCP contributed healthy-adult resting-state scans acquired with multiple
phase-encoding directions \citep{ge2023chcp}.

\paragraph{Human Connectome Project (HCP).}
HCP contributed young-adult resting-state scans to pretraining
\citep{vanessen2013hcp}. It also provided an internal downstream task for sex
classification; the original data partitions remained separated by subject.

\paragraph{Adolescent Brain Cognitive Development (ABCD).}
ABCD contributed pediatric resting-state scans to pretraining
\citep{casey2018abcd}. A separate internal downstream evaluation predicted sex,
with participant identities kept disjoint across the original partitions.

\paragraph{ABCC Data Share.}
ABCC contributed a large pediatric imaging collection
\citep{feczko2021abcc}. Subjects overlapping with ABCD were removed from ABCC
before constructing the pretraining manifest.

\paragraph{Healthy Brain Network (HBN).}
HBN contributed pediatric and adolescent scans spanning a broad range of
mental-health and learning characteristics \citep{alexander2017hbn}.

\paragraph{Philadelphia Neurodevelopmental Cohort (PNC).}
PNC contributed neurodevelopmental resting-state scans from youth and young
adults \citep{satterthwaite2014pnc}.

\paragraph{Major Depressive Disorder (MDD).}
The multisite MDD collection contributed resting-state scans from participants
with recurrent depression and controls \citep{yan2019restmdd}.

\paragraph{Amsterdam Open MRI Collection (AOMIC).}
AOMIC PIOP1 and PIOP2 contributed young-adult resting-state scans
\citep{snoek2021aomic}.

\paragraph{Downstream datasets and tasks.}
Seven datasets provided nine downstream tasks. HCP and ABCD were treated as
in-cohort evaluations because their unlabeled scans contributed to pretraining;
no downstream labels were used. The remaining cohorts were excluded from the
pretraining mixture and provided out-of-cohort transfer evaluations.

\paragraph{Parkinson's Progression Markers Initiative (PPMI).}
PPMI was used for three-class diagnostic-group classification
\citep{marek2011ppmi}.

\paragraph{Alzheimer's Disease Neuroimaging Initiative (ADNI).}
ADNI provided two binary classification tasks: Alzheimer's disease and mild
cognitive impairment \citep{weiner2010adni}.

\paragraph{Autism Brain Imaging Data Exchange (ABIDE).}
ABIDE was used for autism-spectrum-disorder classification
\citep{dimartino2014abide}. ABIDE and ABIDE2 were both excluded from
pretraining.

\paragraph{ADHD-200.}
ADHD-200 provided ADHD diagnosis classification and age prediction
\citep{adhd200consortium2012}. The complete cohort was excluded from
pretraining.

\paragraph{Southwest University Adult Lifespan Dataset (SALD).}
SALD was used for lifespan age prediction and was excluded from pretraining
\citep{wei2018sald}.

For datasets distributed with predefined partitions, we combined the
available samples before constructing participant-grouped cross-validation
folds (Table~\ref{tab:dataset-task-summary}).

\begin{table}[H]
\centering
\caption{\textbf{Downstream evaluation cohorts and prediction targets.}
Participants are grouped across all available segment-level examples
when constructing cross-validation folds.}
\label{tab:dataset-task-summary}
\scriptsize
\renewcommand{\arraystretch}{1.16}
\begin{tabular*}{0.86\linewidth}{@{\extracolsep{\fill}}lrrl@{}}
\toprule
\textbf{Dataset / task} &
\textbf{Participants} &
\textbf{Examples} &
\textbf{Target} \\
\midrule
HCP sex & 1,000 & 5,997 & Binary sex \\
ABCD sex & 500 & 999 & Binary sex \\
PPMI group & 474 & 1,896 & Three diagnostic groups \\
ADNI AD & 230 & 230 & AD vs.\ control \\
ADNI MCI & 292 & 292 & MCI vs.\ control \\
ABIDE ASD & 173 & 173 & ASD vs.\ control \\
ADHD200 diagnosis & 646 & 1,923 & ADHD vs.\ control \\
SALD age & 493 & 493 & Chronological age \\
ADHD200 age & 646 & 1,923 & Chronological age \\
\bottomrule
\end{tabular*}
\end{table}

\subsection{Baseline models}\label{appendix-baseline-models}

The linear-probing comparison used four ROI-based fMRI representation models.
Each encoder was kept frozen, and its scan-level representation was evaluated
with the same task-specific linear-probe family used for CoFlu-Brain. Each
model used its native input parcellation and feature extraction pathway.

\paragraph{BrainGFM.}
BrainGFM pretrains a shared graph encoder across multiple parcellations using
graph contrastive learning and masked graph autoencoding \citep{braingfm}.
We used the authors' released pretrained checkpoint with Schaefer-100 inputs
\citep{schaefer2018local}. Raw Pearson-FC rows served as node features, and
FC values above 0.3 defined the graph adjacency. The encoder's graph-level
output was used as the frozen scan representation.

\paragraph{BrainMASS.}
BrainMASS constructs augmented FC views by dropping BOLD observations and
learns with masked-ROI modeling and latent representation alignment
\citep{brainmass}. In CoFlu-Brain, each partition forms a complementary
retained--dropped FC pair; multiple retained estimates are aligned, and each
paired dropped-view representation is predicted. We used the authors'
released Schaefer-100
checkpoint \citep{schaefer2018local} and released downstream feature path:
Transformer-encoded FC rows followed by dimensionality reduction and
flattening.

\paragraph{BrainLM.}
BrainLM is a masked-autoencoding foundation model for regional fMRI time
series \citep{brainlm}. We used the released 111M-parameter ViT-MAE checkpoint
with its native 424-region input. The concatenation of the final class token
and mean patch representation was used as the frozen scan feature.

\paragraph{Large Connectome Model.}
The Large Connectome Model (LCM) learns connectome representations through
multitask brain--environment interactions \citep{lcm}. We used the released
AAL-116 backbone and decoder checkpoints \citep{tzourio2002aal}, computed raw
Pearson FC following the released input convention, and mean-pooled the final
decoder-query states to obtain a frozen scan representation.

\subsection{Configuration}
\label{appendix-training-and-probing}

CoFlu-Brain was pretrained from scratch using two-rank distributed training.
The model, optimization, and frozen-probe configurations are summarized in
Table~\ref{tab:configuration-summary}.

\begin{table}[H]
\centering
\caption{\textbf{CoFlu-Brain model, pretraining, and linear-probe configurations.}}
\label{tab:configuration-summary}
\scriptsize
\setlength{\tabcolsep}{8pt}
\renewcommand{\arraystretch}{1.08}
\begin{tabularx}{\linewidth}{@{}lX@{}}
\toprule
\textbf{Configuration} & \textbf{Value} \\
\midrule
\multicolumn{2}{@{}l}{\bfseries Model configuration} \\
\multicolumn{2}{@{}l}{\itshape Input representation} \\
ROI atlas & Schaefer-400 \\
Input segment length & 200 frames \\
\multicolumn{2}{@{}l}{\itshape FC-row embedder} \\
MLP dimensions & \(400 \rightarrow 512 \rightarrow 128 \rightarrow 512\) \\
\multicolumn{2}{@{}l}{\itshape ViT encoder} \\
Encoder depth & 6 blocks \\
Embedding dimension & 512 \\
Attention heads & 8 \\
Readout & Mean pooling over ROI tokens \\
\multicolumn{2}{@{}l}{\itshape Projection and prediction heads} \\
Projector MLP dimensions & \(512 \rightarrow 2048 \rightarrow 2048 \rightarrow 128\) \\
Predictor MLP dimensions & \(128 \rightarrow 512 \rightarrow 128\) \\
\addlinespace[2pt]
\multicolumn{2}{@{}l}{\bfseries Pretraining configuration} \\
Optimizer & AdamW \\
Learning rate & \(3\times10^{-4}\) \\
Weight decay & 0.1 \\
Dropout & 0.2 \\
Gradient clipping & 5.0 \\
Total pair batch size & 64 \\
Training epochs & 150 \\
Optimizer updates & 9,600 \\
Temporal partitions & 4 \\
Retained-frame ratio & \(U[0.4,0.7]\) \\
Consistency weight & 0.5 \\
SIGReg weight & 0.1 \\
Consistency schedule & Full weight at epoch 1; linear decay to zero at epoch 25 \\
Checkpoint selection & Last checkpoint \\
\addlinespace[2pt]
\multicolumn{2}{@{}l}{\bfseries Frozen-probe configuration} \\
Encoder & Frozen \\
Representation & Final encoder representation \\
Primary, temporal-sampling, and objective comparison folds & 5 \\
Other sensitivity-analysis folds & 3 \\
Fold construction & Participant-grouped and task-stratified \\
Fold seed & 42 \\
Evaluation cohort & All available downstream samples \\
Five-fold probe fitting & Four folds; evaluation on the remaining fold \\
Three-fold probe fitting & Two folds; evaluation on the remaining fold \\
Feature scaling & StandardScaler fitted on the training folds only \\
Classification model & Logistic regression, \(C=1\) \\
Maximum iterations & 2,000 \\
Classification metrics & Accuracy and weighted F1 \\
Age-regression model & Ridge regression, \(\alpha=1\) \\
Age-target scaling & Training-fold mean and population standard deviation \\
Age-regression metric & \(\mathrm{nMAE}=\mathrm{MAE}/\sigma_{\mathrm{train}}\) \\
Fold summary & Mean \(\pm\) sample SD \\
\bottomrule
\end{tabularx}
\end{table}
\section{Expanded Main Results}\label{appendix-expanded-main-results}

\subsection{Seed-wise linear-probing results}
\label{appendix-pretraining-seed-stability}

Table~\ref{tab:pretraining-seed-stability} reports the five-fold performance
of each independently pretrained CoFlu-Brain encoder. The final column
matches the seed-averaged result in
Table~\ref{tab:existing-model-transfer}.

\begin{table}[!htbp]
\centering
\caption{\textbf{CoFlu-Brain linear probing across pretraining seeds.}
Values are mean \(\pm\) sample SD across five participant-grouped folds.
Metrics are weighted F1 (\(\uparrow\)) and age nMAE (\(\downarrow\)).
The final column averages scores across the three pretraining seeds within each
fold before computing mean \(\pm\) SD.}
\label{tab:pretraining-seed-stability}
\scriptsize
\setlength{\tabcolsep}{4pt}
\renewcommand{\arraystretch}{1.18}
\begin{tabular*}{\linewidth}{@{\extracolsep{\fill}}>{\raggedright\arraybackslash}m{108pt}ccc>{\columncolor{TableTint}}c@{}}
\toprule
\multicolumn{1}{@{}c}{%
\setlength{\unitlength}{1pt}%
\begin{picture}(108,22)
\put(0,22){\line(5,-1){108}}
\put(78,17){\makebox(0,0)[c]{\scriptsize\bfseries Run}}
\put(27,4){\makebox(0,0)[c]{\scriptsize\bfseries Dataset}}
\end{picture}} &
\raisebox{7pt}{\bfseries Seed 42} &
\raisebox{7pt}{\bfseries Seed 43} &
\raisebox{7pt}{\bfseries Seed 44} &
\raisebox{7pt}{\bfseries Seed average} \\
\hline
\bfseries HCP / Sex \(\uparrow\) & 0.7865 \(\pm\) 0.0113 & 0.7648 \(\pm\) 0.0155 & 0.7722 \(\pm\) 0.0142 & 0.7745 \(\pm\) 0.0103 \\
\bfseries ABCD / Sex \(\uparrow\) & 0.5899 \(\pm\) 0.0184 & 0.5828 \(\pm\) 0.0497 & 0.6157 \(\pm\) 0.0182 & 0.5961 \(\pm\) 0.0190 \\
\bfseries SALD / Age \(\downarrow\) & 0.6982 \(\pm\) 0.0586 & 0.6378 \(\pm\) 0.0437 & 0.6353 \(\pm\) 0.0373 & 0.6571 \(\pm\) 0.0431 \\
\bfseries PPMI / Group \(\uparrow\) & 0.5802 \(\pm\) 0.0206 & 0.5785 \(\pm\) 0.0314 & 0.5736 \(\pm\) 0.0187 & 0.5775 \(\pm\) 0.0173 \\
\bfseries ADNI / AD \(\uparrow\) & 0.7219 \(\pm\) 0.0657 & 0.6643 \(\pm\) 0.0425 & 0.6643 \(\pm\) 0.0492 & 0.6835 \(\pm\) 0.0307 \\
\bfseries ADNI / MCI \(\uparrow\) & 0.5461 \(\pm\) 0.0553 & 0.5585 \(\pm\) 0.0573 & 0.5567 \(\pm\) 0.0524 & 0.5538 \(\pm\) 0.0441 \\
\bfseries ABIDE / ASD \(\uparrow\) & 0.6013 \(\pm\) 0.1107 & 0.5623 \(\pm\) 0.0553 & 0.5066 \(\pm\) 0.0621 & 0.5567 \(\pm\) 0.0476 \\
\bfseries ADHD200 / Age \(\downarrow\) & 0.6523 \(\pm\) 0.0315 & 0.6528 \(\pm\) 0.0426 & 0.6620 \(\pm\) 0.0455 & 0.6557 \(\pm\) 0.0370 \\
\bfseries ADHD200 / Diagnosis \(\uparrow\) & 0.5867 \(\pm\) 0.0207 & 0.5836 \(\pm\) 0.0175 & 0.5969 \(\pm\) 0.0394 & 0.5891 \(\pm\) 0.0151 \\
\bottomrule
\end{tabular*}
\end{table}

\subsection{Complete network-restricted probing results}
\label{appendix-complete-network-probing}

Table~\ref{tab:complete-network-probing} expands the top-five presentation in
Figure~\ref{fig:adhd-network-probing} to all 17 Yeo networks. Each value is
the mean and sample SD across the three independently pretrained CoFlu-Brain
encoders. The table reports predictive performance from retaining
subject-specific FC within one network while replacing all remaining entries
with their probe-training mean.

\begin{table}[H]
\centering
\caption{\textbf{Functional-network selection on ADHD200.}
Gray shading marks the top-five networks identified for each task by
network-restricted probing. Values are mean \(\pm\) SD across three independent
pretraining runs. Age prediction reports nMAE (\(\downarrow\)); diagnosis
reports weighted F1 (\(\uparrow\)). ROI counts indicate network size.}
\label{tab:complete-network-probing}
\scriptsize
\setlength{\tabcolsep}{8pt}
\renewcommand{\arraystretch}{1.15}
\newcommand{\networkselected}[1]{%
  \multicolumn{1}{>{\columncolor{TableTint}}r}{\bfseries #1}}
\begin{tabular*}{0.82\linewidth}{@{\extracolsep{\fill}}lcrr@{}}
\toprule
\textbf{Network} & \textbf{ROIs} & \textbf{Age nMAE \(\downarrow\)} & \textbf{Diagnosis F1 \(\uparrow\)} \\
\midrule
DefaultA & 34 & \networkselected{0.6804 \(\pm\) 0.0065} & \networkselected{0.5965 \(\pm\) 0.0165} \\
ContC & 12 & \networkselected{0.6894 \(\pm\) 0.0029} & 0.5619 \(\pm\) 0.0132 \\
LimbicA & 13 & \networkselected{0.6923 \(\pm\) 0.0056} & 0.5442 \(\pm\) 0.0047 \\
DefaultC & 13 & \networkselected{0.7012 \(\pm\) 0.0052} & 0.5288 \(\pm\) 0.0125 \\
SomMotA & 39 & \networkselected{0.7042 \(\pm\) 0.0116} & 0.5435 \(\pm\) 0.0226 \\
LimbicB & 11 & 0.7196 \(\pm\) 0.0138 & 0.5372 \(\pm\) 0.0092 \\
DorsAttnB & 25 & 0.7245 \(\pm\) 0.0099 & 0.5490 \(\pm\) 0.0101 \\
SomMotB & 31 & 0.7262 \(\pm\) 0.0187 & 0.5806 \(\pm\) 0.0153 \\
VisPeri & 23 & 0.7314 \(\pm\) 0.0180 & \networkselected{0.5857 \(\pm\) 0.0032} \\
TempPar & 16 & 0.7446 \(\pm\) 0.0040 & 0.5032 \(\pm\) 0.0179 \\
SalVentAttnA & 34 & 0.7447 \(\pm\) 0.0089 & \networkselected{0.5886 \(\pm\) 0.0192} \\
DefaultB & 32 & 0.7520 \(\pm\) 0.0100 & \networkselected{0.6043 \(\pm\) 0.0117} \\
VisCent & 24 & 0.7526 \(\pm\) 0.0070 & \networkselected{0.5807 \(\pm\) 0.0264} \\
DorsAttnA & 27 & 0.7562 \(\pm\) 0.0110 & 0.4899 \(\pm\) 0.0170 \\
SalVentAttnB & 17 & 0.7912 \(\pm\) 0.0284 & 0.5330 \(\pm\) 0.0115 \\
ContB & 25 & 0.7934 \(\pm\) 0.0020 & 0.5169 \(\pm\) 0.0131 \\
ContA & 24 & 0.7966 \(\pm\) 0.0048 & 0.5157 \(\pm\) 0.0066 \\
\bottomrule
\end{tabular*}
\end{table}

The complete table preserves the patterns summarized in the main text:
DefaultA provided the strongest age-prediction result across all three seeds,
whereas diagnosis was led by DefaultB and followed by DefaultA and
SalVentAttnA.

\section{Ablation and Sensitivity Analyses}
\label{appendix-ablation-and-sensitivity}

\subsection{Random versus ordered frame partitioning}
\label{appendix-temporal-sampling-details}

Table~\ref{tab:temporal-sampling-appendix} reports the task-level results for
random and ordered frame partitioning.

\begin{table}[!htbp]
\centering
\caption{\textbf{Downstream performance under random and ordered frame
partitioning.} Each seed entry is mean \(\pm\) sample SD
across five participant-grouped folds. The final column averages scores across
pretraining seeds within each fold before computing mean \(\pm\) SD. Metrics are weighted
F1 (\(\uparrow\)) and age nMAE (\(\downarrow\)). Red boldface marks the better
seed-averaged score for each task.}
\label{tab:temporal-sampling-appendix}
\scriptsize
\renewcommand{\arraystretch}{1.16}
\textit{\textbf{(a) Random partitioning}}\\[3pt]
\setlength{\tabcolsep}{5pt}
\resizebox{0.94\linewidth}{!}{%
\begin{tabular}{@{}>{\raggedright\arraybackslash}m{94pt}ccc>{\columncolor{TableTint}}c@{}}
\toprule
\multicolumn{1}{@{}c}{%
\setlength{\unitlength}{1pt}%
\begin{picture}(94,22)
\put(0,22){\line(5,-1){94}}
\put(69,17){\makebox(0,0)[c]{\scriptsize\bfseries Run}}
\put(24,4){\makebox(0,0)[c]{\scriptsize\bfseries Dataset}}
\end{picture}} &
\raisebox{7pt}{\bfseries Seed 42} &
\raisebox{7pt}{\bfseries Seed 43} &
\raisebox{7pt}{\bfseries Seed 44} &
\raisebox{7pt}{\bfseries Seed average} \\
\hline
HCP sex \(\uparrow\) & 0.7865 \(\pm\) 0.0113 & 0.7648 \(\pm\) 0.0155 & 0.7722 \(\pm\) 0.0142 & \best{0.7745 \(\pm\) 0.0103} \\
ABCD sex \(\uparrow\) & 0.5899 \(\pm\) 0.0184 & 0.5828 \(\pm\) 0.0497 & 0.6157 \(\pm\) 0.0182 & \best{0.5961 \(\pm\) 0.0190} \\
SALD age \(\downarrow\) & 0.6982 \(\pm\) 0.0586 & 0.6378 \(\pm\) 0.0437 & 0.6353 \(\pm\) 0.0373 & \best{0.6571 \(\pm\) 0.0431} \\
PPMI group \(\uparrow\) & 0.5802 \(\pm\) 0.0206 & 0.5785 \(\pm\) 0.0314 & 0.5736 \(\pm\) 0.0187 & \best{0.5775 \(\pm\) 0.0173} \\
ADNI AD \(\uparrow\) & 0.7219 \(\pm\) 0.0657 & 0.6643 \(\pm\) 0.0425 & 0.6643 \(\pm\) 0.0492 & \best{0.6835 \(\pm\) 0.0307} \\
ADNI MCI \(\uparrow\) & 0.5461 \(\pm\) 0.0553 & 0.5585 \(\pm\) 0.0573 & 0.5567 \(\pm\) 0.0524 & \best{0.5538 \(\pm\) 0.0441} \\
ABIDE ASD \(\uparrow\) & 0.6013 \(\pm\) 0.1107 & 0.5623 \(\pm\) 0.0553 & 0.5066 \(\pm\) 0.0621 & \best{0.5567 \(\pm\) 0.0476} \\
ADHD200 age \(\downarrow\) & 0.6523 \(\pm\) 0.0315 & 0.6528 \(\pm\) 0.0426 & 0.6620 \(\pm\) 0.0455 & \best{0.6557 \(\pm\) 0.0370} \\
ADHD200 diagnosis \(\uparrow\) & 0.5867 \(\pm\) 0.0207 & 0.5836 \(\pm\) 0.0175 & 0.5969 \(\pm\) 0.0394 & \best{0.5891 \(\pm\) 0.0151} \\
\bottomrule
\end{tabular}
}

\vspace{8pt}
\textit{\textbf{(b) Ordered partitioning}}\\[3pt]
\resizebox{0.94\linewidth}{!}{%
\begin{tabular}{@{}>{\raggedright\arraybackslash}m{94pt}ccc>{\columncolor{TableTint}}c@{}}
\toprule
\multicolumn{1}{@{}c}{%
\setlength{\unitlength}{1pt}%
\begin{picture}(94,22)
\put(0,22){\line(5,-1){94}}
\put(69,17){\makebox(0,0)[c]{\scriptsize\bfseries Run}}
\put(24,4){\makebox(0,0)[c]{\scriptsize\bfseries Dataset}}
\end{picture}} &
\raisebox{7pt}{\bfseries Seed 42} &
\raisebox{7pt}{\bfseries Seed 43} &
\raisebox{7pt}{\bfseries Seed 44} &
\raisebox{7pt}{\bfseries Seed average} \\
\hline
HCP sex \(\uparrow\) & 0.7769 \(\pm\) 0.0151 & 0.7585 \(\pm\) 0.0301 & 0.7752 \(\pm\) 0.0103 & 0.7702 \(\pm\) 0.0138 \\
ABCD sex \(\uparrow\) & 0.5817 \(\pm\) 0.0366 & 0.6118 \(\pm\) 0.0402 & 0.5926 \(\pm\) 0.0324 & 0.5954 \(\pm\) 0.0171 \\
SALD age \(\downarrow\) & 0.7936 \(\pm\) 0.0675 & 0.7126 \(\pm\) 0.0689 & 0.7376 \(\pm\) 0.0581 & 0.7480 \(\pm\) 0.0345 \\
PPMI group \(\uparrow\) & 0.5596 \(\pm\) 0.0236 & 0.5614 \(\pm\) 0.0275 & 0.5786 \(\pm\) 0.0293 & 0.5665 \(\pm\) 0.0247 \\
ADNI AD \(\uparrow\) & 0.6467 \(\pm\) 0.0477 & 0.6809 \(\pm\) 0.0815 & 0.6988 \(\pm\) 0.0696 & 0.6755 \(\pm\) 0.0480 \\
ADNI MCI \(\uparrow\) & 0.4627 \(\pm\) 0.0818 & 0.5566 \(\pm\) 0.0443 & 0.5663 \(\pm\) 0.0315 & 0.5285 \(\pm\) 0.0164 \\
ABIDE ASD \(\uparrow\) & 0.4958 \(\pm\) 0.1181 & 0.4996 \(\pm\) 0.1064 & 0.5288 \(\pm\) 0.0583 & 0.5081 \(\pm\) 0.0636 \\
ADHD200 age \(\downarrow\) & 0.6479 \(\pm\) 0.0467 & 0.6662 \(\pm\) 0.0422 & 0.7124 \(\pm\) 0.0455 & 0.6755 \(\pm\) 0.0410 \\
ADHD200 diagnosis \(\uparrow\) & 0.5700 \(\pm\) 0.0226 & 0.5782 \(\pm\) 0.0207 & 0.5557 \(\pm\) 0.0161 & 0.5680 \(\pm\) 0.0103 \\
\bottomrule
\end{tabular}
}
\end{table}

\subsection{Objective components}
\label{appendix-a.-objective-component-ablation-on-representative-tasks}

Table~\ref{tab:objective-ablation-appendix} reports the complete results for
all objective variants.

\begin{table}[!htbp]
\centering
\caption{\textbf{Objective-component ablation across nine downstream tasks.}
Each objective variant was pretrained independently with seed 42 under matched
configurations and is reported as mean \(\pm\) SD across five participant-grouped folds.
Metrics are weighted F1 (\(\uparrow\)) and age nMAE (\(\downarrow\)).
Red boldface marks the best mean for each task.}
\label{tab:objective-ablation-appendix}
\scriptsize
\setlength{\tabcolsep}{4pt}
\renewcommand{\arraystretch}{1.18}
\resizebox{0.98\linewidth}{!}{%
\begin{tabular}{@{}>{\raggedright\arraybackslash}m{96pt}>{\columncolor{TableTint}}cccc@{}}
\toprule
\multicolumn{1}{@{}c}{%
\setlength{\unitlength}{1pt}%
\begin{picture}(96,22)
\put(0,22){\line(5,-1){96}}
\put(67,17){\makebox(0,0)[c]{\scriptsize\bfseries Objective}}
\put(25,4){\makebox(0,0)[c]{\scriptsize\bfseries Dataset}}
\end{picture}} &
\raisebox{7pt}{\bfseries Full} &
\raisebox{7pt}{\bfseries No prediction} &
\raisebox{7pt}{\bfseries No consistency} &
\raisebox{7pt}{\bfseries No SIGReg} \\
\hline
\bfseries HCP / Sex \(\uparrow\) & \best{0.7874 \(\pm\) 0.0115} & 0.6565 \(\pm\) 0.0187 & 0.7789 \(\pm\) 0.0110 & 0.6942 \(\pm\) 0.0153 \\
\bfseries ABCD / Sex \(\uparrow\) & 0.5899 \(\pm\) 0.0184 & 0.5383 \(\pm\) 0.0174 & \best{0.6140 \(\pm\) 0.0260} & 0.5511 \(\pm\) 0.0379 \\
\bfseries PPMI / Group \(\uparrow\) & \best{0.5802 \(\pm\) 0.0206} & 0.5399 \(\pm\) 0.0460 & 0.5576 \(\pm\) 0.0285 & 0.5426 \(\pm\) 0.0720 \\
\bfseries ADNI / AD \(\uparrow\) & \best{0.7219 \(\pm\) 0.0657} & 0.5023 \(\pm\) 0.0850 & 0.6866 \(\pm\) 0.1175 & 0.6065 \(\pm\) 0.0667 \\
\bfseries ADNI / MCI \(\uparrow\) & 0.5461 \(\pm\) 0.0553 & 0.4872 \(\pm\) 0.0901 & \best{0.5501 \(\pm\) 0.0705} & 0.5377 \(\pm\) 0.0410 \\
\bfseries ABIDE / ASD \(\uparrow\) & 0.6013 \(\pm\) 0.1107 & \best{0.6082 \(\pm\) 0.0749} & 0.5994 \(\pm\) 0.0835 & 0.4136 \(\pm\) 0.0587 \\
\bfseries ADHD200 / Diagnosis \(\uparrow\) & \best{0.5872 \(\pm\) 0.0215} & 0.5348 \(\pm\) 0.0309 & 0.5725 \(\pm\) 0.0384 & 0.5643 \(\pm\) 0.0288 \\
\bfseries SALD / Age \(\downarrow\) & 0.6982 \(\pm\) 0.0586 & 0.7674 \(\pm\) 0.0551 & \best{0.6778 \(\pm\) 0.0783} & 0.7625 \(\pm\) 0.0577 \\
\bfseries ADHD200 / Age \(\downarrow\) & \best{0.6523 \(\pm\) 0.0315} & 0.7416 \(\pm\) 0.0567 & 0.6611 \(\pm\) 0.0539 & 0.7132 \(\pm\) 0.0482 \\
\bottomrule
\end{tabular}
}
\end{table}

The complete objective led five of the nine tasks and achieved the highest
mean classification F1 (0.6306). Removing prediction or SIGReg generally
reduced transfer performance, while consistency produced task-dependent
gains. Without SIGReg, representations became highly redundant, with a view
cosine of 0.9999999 and a covariance penalty of 639.19, compared with 8.81
for the complete objective.

\subsection{Pretraining data coverage}
\label{appendix-b.-data-coverage-scaling-details}

Table~\ref{tab:data-coverage-appendix} reports every downstream task for all
data scales; aggregate means are omitted because the task-level pattern is
more informative.

\begin{table}[!htbp]
\centering
\caption{\textbf{Pretraining data coverage and downstream performance across
nine tasks.} Values are mean \(\pm\) SD across probe folds.
Metrics are weighted F1 (\(\uparrow\)) and age nMAE (\(\downarrow\)).
Red boldface marks the best mean for each task.}
\label{tab:data-coverage-appendix}
\scriptsize
\setlength{\tabcolsep}{4pt}
\renewcommand{\arraystretch}{1.18}
\resizebox{0.98\linewidth}{!}{%
\begin{tabular}{@{}>{\raggedright\arraybackslash}m{96pt}ccc>{\columncolor{TableTint}}c@{}}
\toprule
\multicolumn{1}{@{}c}{%
\setlength{\unitlength}{1pt}%
\begin{picture}(96,22)
\put(0,22){\line(5,-1){96}}
\put(68,17){\makebox(0,0)[c]{\scriptsize\bfseries Coverage}}
\put(25,4){\makebox(0,0)[c]{\scriptsize\bfseries Dataset}}
\end{picture}} &
\raisebox{7pt}{\bfseries 12.5\%} &
\raisebox{7pt}{\bfseries 25\%} &
\raisebox{7pt}{\bfseries 50\%} &
\raisebox{7pt}{\bfseries 100\%} \\
\hline
\bfseries HCP / Sex \(\uparrow\) & 0.7265 \(\pm\) 0.0126 & 0.7376 \(\pm\) 0.0117 & 0.7326 \(\pm\) 0.0240 & \best{0.7737 \(\pm\) 0.0092} \\
\bfseries ABCD / Sex \(\uparrow\) & 0.5947 \(\pm\) 0.0266 & \best{0.6153 \(\pm\) 0.0383} & 0.5905 \(\pm\) 0.0332 & 0.5604 \(\pm\) 0.0149 \\
\bfseries PPMI / Group \(\uparrow\) & 0.5204 \(\pm\) 0.0402 & \best{0.5844 \(\pm\) 0.0347} & 0.5625 \(\pm\) 0.0310 & 0.5562 \(\pm\) 0.0582 \\
\bfseries ADNI / AD \(\uparrow\) & 0.6627 \(\pm\) 0.0137 & \best{0.7280 \(\pm\) 0.0348} & 0.6786 \(\pm\) 0.0301 & 0.7235 \(\pm\) 0.0336 \\
\bfseries ADNI / MCI \(\uparrow\) & \best{0.6321 \(\pm\) 0.0267} & 0.5847 \(\pm\) 0.0322 & 0.5884 \(\pm\) 0.0090 & 0.5967 \(\pm\) 0.0422 \\
\bfseries ABIDE / ASD \(\uparrow\) & 0.4414 \(\pm\) 0.0404 & 0.5256 \(\pm\) 0.0614 & 0.5144 \(\pm\) 0.0275 & \best{0.5427 \(\pm\) 0.0500} \\
\bfseries ADHD200 / Diagnosis \(\uparrow\) & 0.5678 \(\pm\) 0.0134 & 0.5609 \(\pm\) 0.0097 & 0.5314 \(\pm\) 0.0333 & \best{0.5756 \(\pm\) 0.0298} \\
\bfseries SALD / Age \(\downarrow\) & 0.9158 \(\pm\) 0.0296 & 0.7858 \(\pm\) 0.0347 & 0.7831 \(\pm\) 0.0281 & \best{0.7610 \(\pm\) 0.0802} \\
\bfseries ADHD200 / Age \(\downarrow\) & 0.7553 \(\pm\) 0.0318 & 0.7453 \(\pm\) 0.0341 & 0.7152 \(\pm\) 0.0174 & \best{0.6785 \(\pm\) 0.0224} \\
\bottomrule
\end{tabular}
}
\end{table}

Increasing data coverage improves both age-regression tasks and five of seven
classification tasks from 12.5\% to 100\%. The individual classification
responses remain nonmonotonic.

\subsection{Number of cofluctuation views}
\label{appendix-c.-branch-count-scaling-details}

Table~\ref{tab:branch-count-appendix} reports every downstream task for all
branch-count variants. The main text reports the four-task summary.

\begin{table}[!htbp]
\centering
\caption{\textbf{Temporal branch-count comparison across nine downstream tasks.}
Values are mean \(\pm\) SD across probe folds.
Metrics are weighted F1 (\(\uparrow\)) and age nMAE (\(\downarrow\)).
Red boldface marks the best mean for each task.}
\label{tab:branch-count-appendix}
\scriptsize
\setlength{\tabcolsep}{4pt}
\renewcommand{\arraystretch}{1.18}
\resizebox{0.98\linewidth}{!}{%
\begin{tabular}{@{}>{\raggedright\arraybackslash}m{96pt}c>{\columncolor{TableTint}}ccc@{}}
\toprule
\multicolumn{1}{@{}c}{%
\setlength{\unitlength}{1pt}%
\begin{picture}(96,22)
\put(0,22){\line(5,-1){96}}
\put(68,17){\makebox(0,0)[c]{\scriptsize\bfseries Branches}}
\put(25,4){\makebox(0,0)[c]{\scriptsize\bfseries Dataset}}
\end{picture}} &
\raisebox{7pt}{\bfseries 2 branches} &
\raisebox{7pt}{\bfseries 4 branches} &
\raisebox{7pt}{\bfseries 6 branches} &
\raisebox{7pt}{\bfseries 8 branches} \\
\hline
\bfseries HCP / Sex \(\uparrow\) & 0.7557 \(\pm\) 0.0205 & \best{0.7737 \(\pm\) 0.0092} & 0.7601 \(\pm\) 0.0296 & 0.7643 \(\pm\) 0.0122 \\
\bfseries ABCD / Sex \(\uparrow\) & 0.5713 \(\pm\) 0.0092 & 0.5604 \(\pm\) 0.0149 & 0.5713 \(\pm\) 0.0444 & \best{0.5895 \(\pm\) 0.0308} \\
\bfseries PPMI / Group \(\uparrow\) & 0.5220 \(\pm\) 0.0446 & 0.5562 \(\pm\) 0.0582 & \best{0.5846 \(\pm\) 0.0373} & 0.5827 \(\pm\) 0.0331 \\
\bfseries ADNI / AD \(\uparrow\) & 0.6518 \(\pm\) 0.0536 & \best{0.7235 \(\pm\) 0.0336} & 0.6429 \(\pm\) 0.0818 & 0.6686 \(\pm\) 0.0470 \\
\bfseries ADNI / MCI \(\uparrow\) & 0.5514 \(\pm\) 0.0207 & \best{0.5967 \(\pm\) 0.0422} & 0.5357 \(\pm\) 0.0371 & 0.5617 \(\pm\) 0.0287 \\
\bfseries ABIDE / ASD \(\uparrow\) & 0.4849 \(\pm\) 0.0917 & \best{0.5427 \(\pm\) 0.0500} & 0.5274 \(\pm\) 0.0394 & 0.4959 \(\pm\) 0.0208 \\
\bfseries ADHD200 / Diagnosis \(\uparrow\) & \best{0.5762 \(\pm\) 0.0387} & 0.5756 \(\pm\) 0.0298 & 0.5637 \(\pm\) 0.0183 & 0.5713 \(\pm\) 0.0269 \\
\bfseries SALD / Age \(\downarrow\) & 0.7605 \(\pm\) 0.0446 & 0.7610 \(\pm\) 0.0802 & \best{0.7139 \(\pm\) 0.0128} & 0.7256 \(\pm\) 0.0274 \\
\bfseries ADHD200 / Age \(\downarrow\) & 0.7088 \(\pm\) 0.0091 & \best{0.6785 \(\pm\) 0.0224} & 0.6872 \(\pm\) 0.0195 & 0.6894 \(\pm\) 0.0338 \\
\bottomrule
\end{tabular}
}
\end{table}

Pretraining cost increased from 10.54 GPU-hours for four branches to 26.72
GPU-hours for eight branches. Performance varied nonmonotonically across
tasks, with no consistent benefit from increasing the branch count beyond
four. All runs completed the same 9,600-update budget.

\subsection{Consistency-weight schedule}
\label{appendix-d.-loss-weight-and-schedule-sensitivity}

Table~\ref{tab:loss-schedule-appendix} expands the three schedules compared in
the main text to all nine downstream tasks.

\begin{table}[H]
\centering
\caption{\textbf{Consistency-weight schedule comparison across nine downstream tasks.}
Values are mean \(\pm\) SD across probe folds.
Metrics are weighted F1 (\(\uparrow\)) and age nMAE (\(\downarrow\)).
Red boldface marks the best mean for each task.}
\label{tab:loss-schedule-appendix}
\scriptsize
\setlength{\tabcolsep}{6pt}
\renewcommand{\arraystretch}{1.14}
\resizebox{0.84\linewidth}{!}{%
\begin{tabular}{@{}>{\raggedright\arraybackslash}m{104pt}>{\columncolor{TableTint}}ccc@{}}
\toprule
\multicolumn{1}{@{}c}{%
\setlength{\unitlength}{1pt}%
\begin{picture}(104,22)
\put(0,22){\line(5,-1){104}}
\put(74,17){\makebox(0,0)[c]{\scriptsize\bfseries Schedule}}
\put(27,4){\makebox(0,0)[c]{\scriptsize\bfseries Dataset}}
\end{picture}} &
\raisebox{7pt}{\bfseries Decay-25} &
\raisebox{7pt}{\bfseries Decay-50} &
\raisebox{7pt}{\bfseries Constant} \\
\hline
\bfseries HCP / Sex \(\uparrow\) & \best{0.7737 \(\pm\) 0.0092} & 0.7588 \(\pm\) 0.0204 & 0.7687 \(\pm\) 0.0164 \\
\bfseries ABCD / Sex \(\uparrow\) & 0.5604 \(\pm\) 0.0149 & \best{0.5955 \(\pm\) 0.0349} & 0.5916 \(\pm\) 0.0442 \\
\bfseries PPMI / Group \(\uparrow\) & \best{0.5562 \(\pm\) 0.0582} & 0.5179 \(\pm\) 0.0120 & 0.5540 \(\pm\) 0.0048 \\
\bfseries ADNI / AD \(\uparrow\) & \best{0.7235 \(\pm\) 0.0336} & 0.6736 \(\pm\) 0.0236 & 0.6627 \(\pm\) 0.0350 \\
\bfseries ADNI / MCI \(\uparrow\) & \best{0.5967 \(\pm\) 0.0422} & 0.5636 \(\pm\) 0.0107 & 0.5431 \(\pm\) 0.0253 \\
\bfseries ABIDE / ASD \(\uparrow\) & 0.5427 \(\pm\) 0.0500 & 0.4734 \(\pm\) 0.0243 & \best{0.5669 \(\pm\) 0.0179} \\
\bfseries ADHD200 / Diagnosis \(\uparrow\) & \best{0.5756 \(\pm\) 0.0298} & 0.5433 \(\pm\) 0.0282 & 0.5688 \(\pm\) 0.0300 \\
\bfseries SALD / Age \(\downarrow\) & 0.7610 \(\pm\) 0.0802 & 0.7607 \(\pm\) 0.0261 & \best{0.7576 \(\pm\) 0.0617} \\
\bfseries ADHD200 / Age \(\downarrow\) & 0.6785 \(\pm\) 0.0224 & 0.6839 \(\pm\) 0.0239 & \best{0.6568 \(\pm\) 0.0270} \\
\bottomrule
\end{tabular}
}
\end{table}

Decay-25 led five of the seven classification tasks, Decay-50 led ABCD sex,
and Constant led ABIDE ASD and both age-prediction tasks. The complete results
preserve the task-dependent schedule pattern summarized in the main text.

\end{document}